\documentclass{article}
\usepackage{iclr2027_conference,times}

\usepackage{amsmath,amsfonts,bm}

\def\eqref#1{equation~\ref{#1}}

\def\1{\bm{1}}

\DeclareMathAlphabet{\mathsfit}{\encodingdefault}{\sfdefault}{m}{sl}
\SetMathAlphabet{\mathsfit}{bold}{\encodingdefault}{\sfdefault}{bx}{n}

\usepackage{hyperref}
\usepackage{url}
\usepackage{graphicx} 
\usepackage{booktabs}
\usepackage{mathrsfs}
\usepackage{longtable}
\usepackage[margin=1in]{geometry}
\usepackage{multirow}
\usepackage{siunitx}
\usepackage{xcolor}
\usepackage{caption}
\usepackage{subcaption}
\usepackage{colortbl}
\usepackage{makecell}
\usepackage{array}
\usepackage{wrapfig}
\usepackage{tikz} \usepackage{xcolor} \usepackage{fontawesome5} 

\title{Mol-JEPA: A multimodal Joint Embedding Predictive Architecture for Molecules}

\author{%
  \makebox[\linewidth][c]{\textbf{Florian Rottach$^{*1,2}$ \quad Sebastian Schieferdecker$^{*2}$ \quad William Rudman$^{3}$}} \\[0.3em]
  \makebox[\linewidth][c]{\textbf{Randall Balestriero$^{4}$ \quad Carsten Eickhoff$^{1}$}} \\[0.6em]
  \makebox[\linewidth][c]{$^1$University of Tübingen \quad $^2$Boehringer Ingelheim} \\
  \makebox[\linewidth][c]{$^3$The University of Texas at Austin \quad $^4$Brown University} \\[0.3em]
  \makebox[\linewidth][c]{\footnotesize{\texttt{florian.rottach@boehringer-ingelheim.com}}} \\[0.2em]
  \makebox[\linewidth][c]{\scriptsize $^*$Equal contribution} 
}

\definecolor{groupbg}{gray}{0.93}
\definecolor{groupbg}{RGB}{240,240,240}
\definecolor{molbg}{RGB}{235,245,255}
\definecolor{good}{RGB}{0,120,0}
\definecolor{bad}{RGB}{180,0,0}

\definecolor{slategray}{RGB}{226, 232, 240} 
\definecolor{moljepa}{RGB}{235, 248, 255}   
\definecolor{stdgrey}{RGB}{113, 128, 150}   
\usepackage[title,titletoc]{appendix} 
\usepackage{titletoc}      

\newcommand{\std}[1]{\textcolor{stdgrey}{\tiny\,$\pm$\,#1}}

\iclrpreprintcopy
\begin{document}
\maketitle

\begin{abstract}
Despite recent advances in molecular foundation models, several limitations remain, such as chemically invalid augmentations, modality collapse, and incomplete representation of biochemical environments. To address these challenges, we present \textbf{Mol-JEPA}, a scalable framework for learning molecular world models. Rather than relying on suboptimal molecular perturbations, our model uses modality masking to exploit information from molecular structures, cellular phenotypes, binding affinities, ADMET profiles, quantum chemistry simulations and other drug discovery data. Across various benchmarks, we show that the representations learned by Mol-JEPA deliver strong performance, demonstrating the value of incorporating biochemical context through latent space prediction. 
\end{abstract}

\begin{figure}[h]
    \centering
    \includegraphics[width=0.6\linewidth]{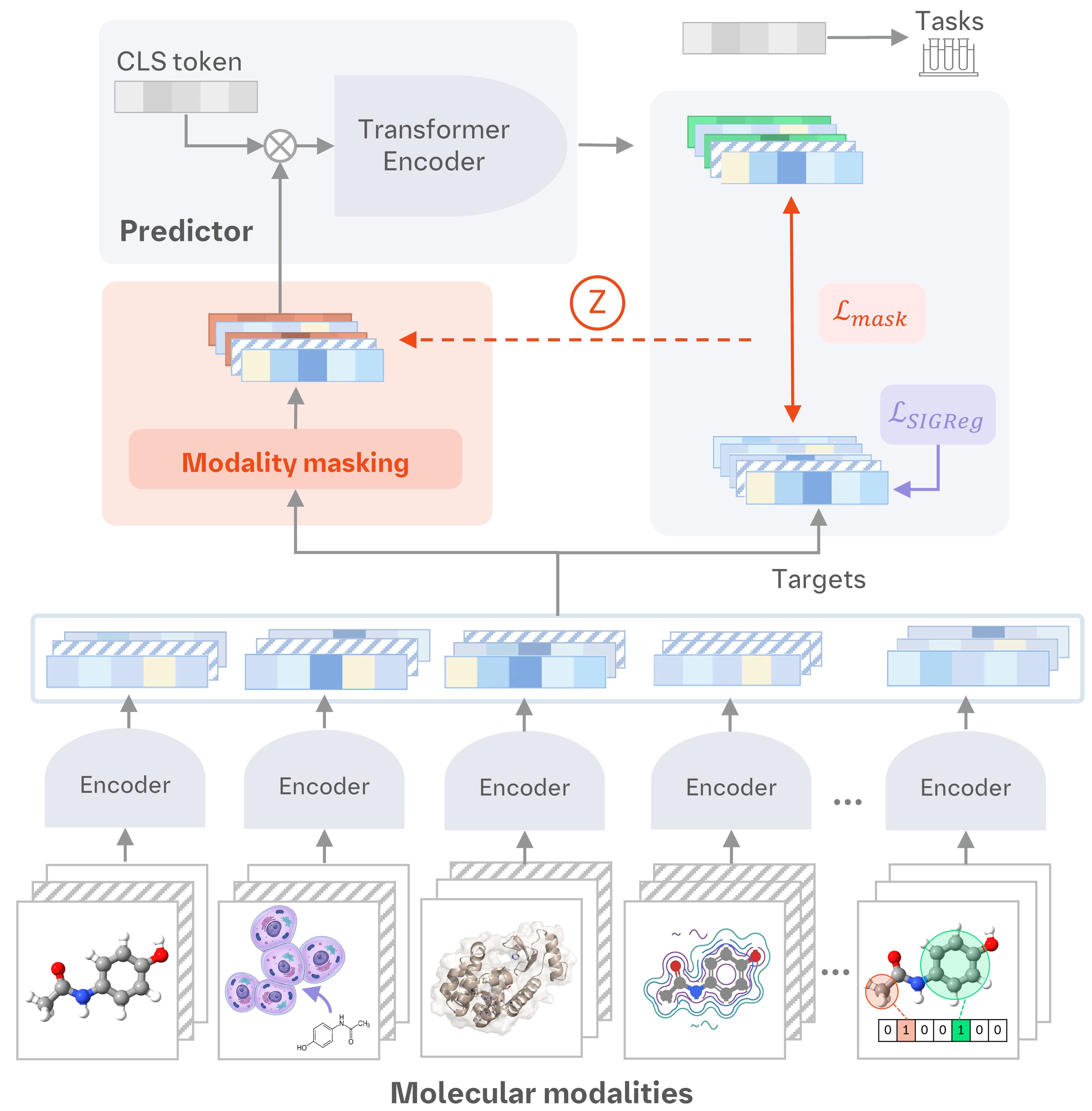}
    \caption{\textbf{Mol-JEPA:} Rather than representing molecules solely through their structure, our model learns from a spectrum of biochemical and physical modalities. During training, modalities are randomly masked and predicted from the remaining inputs using a Transformer-based predictor. To prevent representation collapse, we employ isotropic regularization through SIGReg. }
    \label{fig:overview}
\end{figure}

\section{Introduction}
The development of novel drugs is a costly process characterized by high failure rates. Drug candidates may fail for numerous reasons, including weak target binding, poor absorption, rapid metabolic clearance, and adverse toxicity profiles \cite{sun202290}. Over the past decades, cheminformatics research has led to the development of various molecular representations to address the challenging prediction problems in drug discovery. Examples range from basic physicochemical descriptors, such as atom and bond statistics to more advanced features including Extended Connectivity Fingerprints (ECFP) \cite{rogers2010extended} and quantum-chemical descriptors \cite{wang2021quantum}. In recent years, learned representations have emerged from various neural network variants, such as Graph Neural Networks \cite{reiser2022graph}, which operate directly on the molecular graph or Chemical Language Models \cite{grisoni2023chemical}, representing molecules as text. While these works are important milestones, several studies have found that pretrained representations often fail to consistently outperform traditional baselines or only show marginal performance gains \cite{jiang2021could, sun2022does, fischer2025deep, praski2025benchmarking, guo2026larger}. On one hand, this can be attributed to the pretraining datasets, which either have poor quality or do not cover the relevant molecular space \cite{rodrigues2019good, chodera2026blind}. On the other hand, models trained with self-supervised learning (SSL) methods use augmentation strategies that are unsuitable for molecular data. For example, many SSL approaches generate multiple views of a molecule through atom or bond masking and train the model to produce similar representations for these augmented views. This can be problematic because molecular property landscapes are often highly discontinuous, with small structural modifications leading to substantial changes in molecular properties, as demonstrated by property cliffs \cite{stumpfe2019evolving}. This misalignment between augmentation bias and desired inductive invariance has been shown to harm SSL performance \cite{van2025joint} and offers an explanation why SSL approaches on molecular data have not been particularly successful. 

In addition to unsuitable augmentation strategies, we argue that molecular structure alone is insufficient for SSL pretraining, because molecules are inherently \textit{context-dependent}. Drug behavior emerges from complex interactions with biological systems, including protein networks, metabolic processes, cellular states, and physiological environments that are not encoded in molecular structure alone. Incorporating data that reflects molecular interactions enables models to navigate chemical space through a biological lens and may improve their ability to identify complex structure-property relationships, including property cliffs \cite{dablander2023exploring, sanchez2026large}. In recent years, several attempts have been made to enrich molecular representations by incorporating additional information. Some of these models rely on multi-task pretraining \cite{beaini2023towards}, which however suffers from over-specialization due to discontinuities and errors in the data \cite{sheridan2020experimental}. Joint embedding approaches have been shown to be more robust to variations in the input and target spaces \cite{van2025joint}, making them particularly interesting for molecular data. Several works have adapted molecular joint embedding models using contrastive learning \cite{liang2022mind, masood2026unifying, xiong2026multi}. While effective, these methods depend heavily on the selection of negative samples, which can substantially affect representation quality \cite{jing2021understanding}. The recently presented Joint Embedding Predictive Architecture (JEPA) \cite{lecun2022path} has several advantages over previous methods, such as no need for negative sampling and robustness to noise, making it less prone to overfitting. However, JEPA still relies on meaningful data augmentations, which remain challenging to define for molecular structures.

To address these limitations, we present Mol-JEPA, a multimodal molecular world model based on the Joint Embedding Predictive Architecture. Rather than applying augmentations to the molecular structure, we mask entire modalities and predict the corresponding latent representations, enabling the model to learn predictive representations that capture molecular behavior across diverse biological and chemical contexts. Our considered modalities are derived from pretrained foundation models, quantum-chemical calculations and experimental datasets, spanning binding affinity, cellular effects and ADMET properties. Furthermore, by using the LeJEPA framework \cite{balestriero2025lejepa} as anti-collapse mechanism, we ensure that each modality contributes during training. Across relevant downstream datasets, we show that our model learns effective molecular representations outperforming our baselines and reaching strong performance on public leaderboards. Overall, our results demonstrate the potential of latent-space molecular foundation models and establishes a scalable framework for integrating additional modalities and datasets. Our main contributions are as follows:

\begin{itemize}
    \item We present \textbf{Mol-JEPA}, a scalable multimodal Joint Embedding Predictive Architecture that enables self-supervised learning through meaningful augmentations on molecular data in drug discovery.
    \item We introduce a novel multimodal dataset containing nearly 5 million molecules and spanning a diverse drug-like chemical space across a range of biological, chemical, and computational modalities.
    \item We demonstrate highly competitive performance on relevant benchmark datasets, statistically evaluate the quality of predictions, assess out-of-distribution performance and analyze the contributions of different modalities.
\end{itemize}

\section{Related Work}

\paragraph{Deep learning in chemistry}
In recent years, various deep learning methods have been applied in computational chemistry. Early works adapted popular neural network architectures to molecular data, yielding models such as ChemBERTa \cite{chithrananda2020chemberta}, Chemical VAE \cite{gomez2018automatic} or Chemformer \cite{irwin2022chemformer}. Graph Neural Networks (GNNs) \cite{kipf2016semi} have become a prominent approach for molecular representation learning, as they integrate inductive priors that allow them to operate directly on molecular graphs. Others have represented molecules as text for training Chemical Language Models \cite{grisoni2023chemical}, most commonly using the SMILES notation \cite{weininger1988smiles}. While these approaches demonstrated the ability to learn representations directly from molecular data, they remain limited to string-based or graph-based inputs. 
To address this limitation, subsequent work incorporated geometric information such as interatomic distances in GEM \cite{fang2022geometry} and atomic coordinates in Uni-Mol \cite{zhou2023uni}, using a roto-translation equivariant transformer architecture. Following the trends in other domains, pretraining efforts have been scaled to increasingly large datasets, such as UMA \cite{wood2026family}, an interatomic potential model trained on more than 500 million three-dimensional atomic structures.

\paragraph{Self-supervised molecular representation learning}
Popular self-supervised learning paradigms, such as masked prediction and contrastive learning have also found their way into the chemical sciences. For example, several methods have generated different views on a molecule by augmenting molecular graphs, which are jointly encoded or predicted \cite{rong2020self, hafidi2020graphcl, wang2022molecular, mendez2024mole}. We emphasize that such augmentations can introduce false positives; it cannot be guaranteed that two similar views \textit{actually} behave similarly on a specific downstream tasks. Other works have adopted CLIP \cite{radford2021learning} to fuse molecular graphs with other modalities, such as jointly encoding molecules with cellular images \cite{sanchez2023cloome, masood2026unifying} or images of molecules \cite{harnik2025data}. These models overcome the augmentation bias of previous methods, but require negative samples, which can also introduce false negatives. Joint Embedding Predictive Architectures have been applied to augmented views of molecules \cite{mizera2024graph, piccoli2026joint, chemjepa2025, iyer2026m}, which avoid negative samples, but suffer from the same augmentation bias as the previous methods. Recent work introduced CheMeleon \cite{green2026deep}, a GNN pretrained to predict molecular descriptors. While this is a promising pretext task, their model only considers descriptors and does not readily scale to multimodal settings.

\paragraph{Multimodal architectures}
While several multimodal models integrate molecular graphs, SMILES, textual descriptions, and 3D structural information \cite{mirza2024bridging, luo2023molfm, manolache2024molmix}, most focus solely on representations of the molecule itself, neglecting modalities that reflect its effects in biochemical environments. Furthermore, existing approaches typically fuse modalities using concatenation or attention-based aggregation, which can result in modality collapse, causing some modalities to be underutilized or ignored entirely. Joint embedding approaches offer an alternative that avoids these issues. For example, BioXMol aligns cellular, genetic, and molecular modalities within a shared embedding space encouraging the model to leverage information from all modalities during training \cite{masood2026unifying}. While this work offers a promising direction, it relies on contrastive methods, which are known to suffer from modality gap and hubness effects, arising from its discriminative objective \cite{liang2022mind}. Additionally, such methods do not scale well to many modalities and struggle with sparse and only partially paired modalities. 

Our architecture addresses all limitations of previous approaches. First, no chemically invalid molecule augmentations are introduced - instead we mask entire modalities. Second, we go beyond molecular structure by integrating knowledge about the biochemical environment. Third, all modalities are predicted, which avoids modality collapse and encourages the model to leverage all available information. Finally, the framework scales well to multiple modalities, is easy to optimize and does not require any heuristics, such as alternating gradients, stop-loss or student-teacher architectures, commonly found in other works. Overall, we find that the combination of multimodality and predictions in the latent space, both theoretically and empirically prove to be a promising direction towards molecular world models.

\section{Method}
\label{sec:method}

\subsection{Pretraining Data}
We construct a pretraining dataset by merging and curating public datasets and augment them with additional modalities ranging from specialized model embeddings to quantum chemical calculations. Furthermore, we leverage experimental measurements to construct multidimensional molecular profiles that capture diverse aspects of a molecule's chemical and biological behavior. Table \ref{datasets} provides a summary of the individual datasets, which result in a combined dataset size of 4.66 million unique compounds. 

\paragraph{Public datasets}
First, we extract a subset of drug-like small molecules from ChEMBL \cite{gaulton2017chembl} and query the API to add bioactivity information for each molecule. For potency, we filter the activity types to fall into $\text{IC}_{50}$, $\text{K}_{\text{i}}$, $\text{K}_{\text{d}}$, $\text{EC}_{50}$ or $\text{AC}_{50}$ and use the pChEMBL value (-log10 M) for a consistent scale. For absorption, distribution, metabolism, excretion, and toxicity (ADMET) properties, we standardize units and apply further processing steps, as detailed in Appendix \ref{a:datasets}. We drop all activities with less than 30 labels, leaving us with 306 unique assays for around 30 thousand molecules. In addition, we integrate 21 ADME, 11 toxicity and 12 high-throughput screening endpoints from Therapeutic Data Commons (TDC) \cite{huang2021therapeutics}. We aggregate all measurements per molecule and obtain 672 attributes, with on average 8 available labels per molecule. A full table of properties is provided in Appendix Table \ref{tab:tdc_summary}. We also use PCBA-1328 \cite{beaini2023towards}, a subset of PubChem \cite{kim2016pubchem}, containing 1,328 bioassays with "Active" or "Inactive" flags with at least 10 labels per assay. Further, we integrate quantum-chemical data from $\nabla^2$DFT \cite{khrabrov2024nabla}, containing 17 global density functional theory (DFT) properties, such as total energy, dipole moments and HOMO-LUMO gap, computed at the $\omega$B97X-D/def2-SVP theory level. We extract the lowest energy conformer and assign it to each unique SMILES string. All DFT properties have been normalized to yield smooth label distributions, by applying scikit-learn's quantile transformer \cite{kramer2016scikit}. Finally, to increase the chemical space coverage relevant for drug discovery, we sample around 2 million unlabeled molecules from the Enamine REAL space \cite{Grygorenko2020}. We compute international chemical identifiers (InChI), which are used to merge and de-duplicate the dataset, leaving us with a final size of 4.66 million compounds. All molecule SMILES are canonicalized and normalized to pH7 with MoKa (version 3.2.3), stereochemistry is validated and a low energy conformer is computed for each molecule using the ETKDGv3 algorithm \cite{wang2020improving} followed by geometry optimization with the universal force field (UFF) \cite{rappe1992uff}.

\begin{table}[tbp]
  \centering
  \caption{Pretraining datasets with their final sample counts and average molecular weight (MW).}
  \label{datasets}
  \setlength{\tabcolsep}{8pt}
  \renewcommand{\arraystretch}{1.15}
  \small
  \begin{tabular}{llrrrr}
    \toprule
    \textbf{Dataset} & \textbf{Domain} & \textbf{\# Raw} & \textbf{\# Processed} & \textbf{Avg. MW} & \textbf{Source} \\
    \midrule
    ChEMBL        & Bioactivity   & 280,271   & 268,206   & 399.73 & \cite{gaulton2017chembl}   \\
    TDC           & ADMET         & 512,303   & 402,881   & 365.86 & \cite{huang2021therapeutics}   \\
    PCBA-1328     & Bioactivity   & 1,563,664 & 1,124,981 & 378.82 & \cite{beaini2023towards}   \\
    $\nabla^2$DFT & Quantum chemistry & 1,936,931 & 1,270,806 & 307.46 & \cite{khrabrov2024nabla}   \\
    Enamine REAL  & Unlabeled     & 2,014,554 & 2,012,628 & 307.13 & \cite{Grygorenko2020}   \\
    \bottomrule
  \end{tabular}
\end{table}

\paragraph{Computed data}
Our model architecture allows us to distill information from other chemical foundation models, by predicting their learned representations. To this end, we extract final-layer embeddings from several pretrained molecular backbones. First, we compute embeddings for UMA, a large mixture-of-linear-experts graph network trained on DFT simulation data \cite{wood2026family}. Due to computational constraints, we use the smallest model variant, \textit{uma-s-1p2}. We further incorporate CLOOME \cite{sanchez2023cloome}, which jointly encodes molecular structures and cell painting data, offering a deep phenotypic profile that explains the cellular bioactivity of a molecule. Similarly, we integrate BioXMol \cite{masood2026unifying}, a multimodal model trained on cell painting and transcriptomics data using contrastive learning learning. As a general-purpose molecular language model, we additionally extract embeddings from ChemGPT (4.7M) \cite{frey2023neural}, which was pretrained on the PubChem10M dataset. Recent works emphasize that prioritizing binding affinity has the potential to lead to more successful drug candidates and better pharmacological understanding \cite{murcko2026affinity}. Therefore, we attempt to guide the model about adverse effects to improve downstream performance on ADMET properties, such as \textit{in vivo} phenotypes. For this, we perform co-folding and affinity prediction using Boltz-2 \cite{passaro2025boltz} and concatenate the predictions with the ensemble embeddings from the final layer of the binding affinity module. Specifically, we predict against targets from the Bowes-44 panel \cite{Bowes2012}, which are associated with adverse effects based on historical clinical failures. Due to substantial computational requirements, we restrict the predictions to a subset of 9 targets: 4 G-protein coupled receptors (serotonin 5-HT$_{2A}$, dopamine D$_{2}$, muscarinic M$_{2}$, $\alpha_{1}$-adrenoceptor), 2 ion channels (hERG, NMDA GluN1), 1 kinase (Lck), 1 enzyme (COX-1) and 1 nuclear receptor (GR). Predictions were generated for more than 100,000 randomly selected molecules using eight RTX A6000 GPUs with cached multiple-sequence alignments.

In addition to the embeddings extracted from pretrained models, we computed quantum chemical properties using GFN2-xTB \cite{Bannwarth2019}. We choose this semi-empirical method due to its balance of accuracy and performance, allowing us to compute descriptive vectors for a large number of samples. Lastly, we compute descriptors using the Molecular Operating Environment (MOE) \cite{MOE2022} and Extended Connectivity Fingerprints \cite{rogers2010extended}, which are commonly used in computational chemistry. This allows us to integrate human expertise, built up over decades of drug discovery, integrating various relevant properties such as relevant substructures, charge distributions and 3D topology. 

\subsection{Modality encoders}
The combined dataset results in different \textit{views} on a molecule, serving as distinct modalities in our model. Notably, we can treat multi-dimensional label vectors as modalities, as they describe how molecules behave in the chemical world. Combined with the embeddings and descriptors, we obtain 14 final modalities, which are summarized in Table \ref{modalities}. We employ three learnable encoders that project the modalities to the same dimensionality: An atom encoder, designed for atomic coordinates of the shape \textit{(batch, \#atoms, dim)}, a Graph encoder operating on molecular graphs with additional connectivity information and a Vector encoder, which transforms representations of shape \textit{(batch, dim)}.

\begin{table}[tbp]
\centering
\caption{Molecular modalities used in Mol-JEPA.}
\label{modalities}
\footnotesize 
\renewcommand{\arraystretch}{1.1}
\setlength{\tabcolsep}{3.5pt} 
\begin{tabular}{l l r l l r r}
\toprule
\textbf{Modality} & \textbf{Description} & \textbf{Dim} & \textbf{Type} & \textbf{Encoder} & \textbf{\# Samples} & \textbf{Source} \\
\midrule
UMA       & Atomistic foundation model & 128  & Embedding & Atom   & 4,663,778 & \cite{wood2026family} \\
ChemGPT   & SMILES Transformer & 2048 & Embedding & Vector & 4,688,836 & \cite{frey2023neural} \\
CLOOME    & Cell painting model & 512 & Embedding & Vector & 4,640,188 & \cite{sanchez2023cloome} \\
BioXMol   & Multi-assay phenotypic model & 1024 & Embedding & Vector & 4,640,188 & \cite{masood2026unifying} \\
Boltz-2   & Boltz-2 embeddings & 6912 & Embedding & Vector & 102,234 & \cite{passaro2025boltz} \\
Boltz-2 (P) & Boltz-2 binding affinity & 4 & Embedding & Vector & 102,234 & \cite{passaro2025boltz} \\
GNN       & Graph transformer model & 512 & Embedding & Graph & 4,693,006 & \cite{shi2020masked} \\
\midrule
MOE       & Molecular descriptors & 227 & Descriptors & Vector & 4,663,780 & \citeauthor{MOE2022} \\
ECFP      & Circular topology fingerprints & 2048 & Descriptors & Vector & 4,663,780 & \cite{rogers2010extended} \\
xTB       & GFN2-xTB simulations & 7 & Descriptors & Vector & 3,435,704 & \cite{Bannwarth2019} \\
$\nabla^2$DFT & $\omega$B97X-D/def2-SVP DFT  & 17 & Descriptors & Vector & 1,270,722 & \cite{khrabrov2024nabla} \\
\midrule
ChEMBL    & Bioactivity measurements & 306 & Experimental & Vector & 268,188 & \cite{gaulton2017chembl} \\
PCBA      & Bioactivity measurements & 1328 & Experimental & Vector & 1,124,829 & \cite{beaini2023towards} \\
TDC       & ADMET measurements & 672 & Experimental & Vector & 402,755 & \cite{huang2021therapeutics} \\
\bottomrule
\end{tabular}
\end{table}

\subsection{Model Architecture}

\textbf{Mol-JEPA} is a multimodal joint embedding predictive architecture, which predicts masked modalities from available ones (see Figure \ref{fig:overview}). Given a masking ratio $r$, some of the encoded modalities are deactivated, while ensuring that always at least one modality is available. In the masked forward pass, we replace these embeddings by masking tokens and also replace all missing modalities through respective tokens. The prediction module is implemented as a transformer encoder and optimized to predict the masked embeddings. We use the \textit{TransformerEncoderLayer} from PyTorch \cite{paszke2019pytorch}, which applies self-attention and a feedforward network. Further, we add a learnable CLS token as a global readout for downstream tasks and apply positional encodings to all predictor inputs, to track which embeddings need to be predicted. We experiment both with predicting the embeddings directly from the transformer (modality-based) and predicting them from the CLS token through another transformation (CLS-based). To avoid mode collapse, we apply Sketched Isotropic Gaussian Regularization (SIGReg) \cite{balestriero2025lejepa}. We consider multiple strategies for SIGReg, applying it either to the target embeddings, the predicted embeddings or the CLS token. Mol-JEPA is implemented using the stable-pretraining framework \cite{balestriero2025stable} and hyperparameter-tuned with Optuna \cite{akiba2019optuna} over the search space specified in Appendix \ref{a:implementation}, resulting in a final model with around 50 million parameters.

\subsection{Pretraining Objectives}
Assume a batch $\mathcal{B}=\{x_1,\ldots,x_B\}$ with a set of modalities $\mathcal{M}=\{1,\ldots,M\}$. For each sample $i$ and modality $m\in\mathcal M$, an embedding $\mathbf{s}_i^m \in \mathbb{R}^d$ is obtained using encoder $e^m$. Further, let $\mathbf{a}_{i}=(a^1_{i},\ldots,a^M_{i})$, $a^m_{i}\in\{0,1\}$ denote the modality-availability mask for each sample $i$, where $a^m_{i}=1$ if modality $m$ is present. 

For each $i$, we sample a binary mask vector using masking ratio $r$, which determines the modalities to predict:
\[ 
\mathbf{z}_i=(z^1_{i},\ldots,z^M_{i}), \qquad z^m_{i} \sim \mathrm{Bernoulli}(r\,a^m_{i}) 
\]

We ensure that at least one modality remains available after masking and denote the masked modalities $m$ for sample $i$ with $\hat{\mathbf{s}}^m_{i}$. A Transformer predictor is then trained to reconstruct the embeddings of the masked modalities from the available context. This reconstruction task encourages the model to combine information from different modalities and reason about how samples are represented across alternative environments. The prediction error is implemented as the mean squared error between the masked modalities $\hat{\mathbf{s}}^m_{i}$ and the predicted modalities $\hat{\mathbf{s}}^m_{i, \mathrm{pred}}$, across all samples in the batch of size $B$. For each modality $m \in \mathcal{M}$ we have:

\begin{equation} \mathcal{L}^m_{\text{pred}} = \frac{1}{B} \sum_{i=1}^{B} \left\| \hat{\mathbf{s}}^m_{i} - \hat{\mathbf{s}}^m_{i,\mathrm{pred}} \right\|_2^2 \end{equation}

To avoid the representations collapsing to a trivial subspace, Sketched Isotropic Gaussian Regularization (SIGReg) conditions the latent variables to approximate an isotropic distribution $\mathcal{N}(0, I_K)$. SIGReg computes this efficiently using randomized sketching. Given a batch of $B$ embeddings $\{s_i^m\}_{i=1}^{B}$, the method projects them onto $L$ random 1D directions $\{a_l\}_{l=1}^{L}$ drawn uniformly from the unit sphere $\mathbb{S}^{K-1}$. The Epps-Pulley statistic $T$ is then applied to each 1D slice to measure the divergence from a standard 1D normal distribution. The total SIGReg loss is computed as the average of these univariate tests over all random projections:

\begin{equation} \mathcal{L}^m_{\text{SIGReg}} = \frac{1}{L} \sum_{l=1}^{L} T\!\left(\{a_l^\top s_i^m\}_{i=1}^{B}\right) \end{equation}

For our experiments, we fix the number of random projections to 1024, the range of frequencies in Epps-Pulley to $T_{\max}=3.0$, and the number of evaluated points to 17. The final Mol-JEPA objective combines the modality prediction and isotropic regularization losses across all modalities, balanced by $\lambda$: 

\begin{equation} 
\mathcal{L}_{\text{JEPA}} = \sum_{m \in \mathcal{M}} \left( \mathcal{L}^{m}_{\text{pred}} + \lambda\,\mathcal{L}^{m}_{\text{SIGReg}} \right), \qquad \lambda > 0 
\end{equation}

\subsection{Benchmark Datasets}
Traditional molecular benchmarks have been criticized for limited real-world relevance and data quality issues, including noisy labels, inconsistent chemical representations, and undefined stereochemistry \cite{walters2023benchmarks, liu2024welqrate}. To ensure a realistic evaluation, we compare the model predictions across data from recent OpenADMET blind challenges along with a diverse collection of other high quality datasets. First, we use 9 endpoints from the OpenADMET ExpansionRx competition, which contain entirely novel compounds \cite{castellanoslessons}. Similarly, we evaluate on 7 datasets from the ASAP-Polaris challenge \cite{macdermott2026computational}, including binding against two CoVID targets and multiple ADMET endpoints. Additionally, we evaluate on data from the OpenADMET Pregance-X Receptor (PXR) activity prediction blind challenge \cite{fraser2026mapping}. Lastly, we include Biogen ADME \cite{fang2023prospective}, which contains six high quality \textit{in vitro} endpoints measured under the same experimental conditions. We construct 3 test-train splits using Taylor-Butina Clustering \cite{butina1999unsupervised} on 1024 bit ECFP4 fingerprints, by grouping compounds based on a Tanimoto similarity threshold of 0.65 and assign entire clusters to either the training or test set. Furthermore, for the datasets where official competition splits are available, we report performance on these specific subsets. In Appendix \ref{a:datasets} we report additional information about the benchmark datasets and splits.

\subsection{Baselines}
We compare the performance of our model with several established state-of-the-art baselines. As classical approaches, we fit a Random Forest and a Light Gradient Boosting Machine using scikit-learn \cite{pedregosa2011scikit} and LightGBM \cite{ke2017lightgbm} respectively. We experiment with ECFP4 \cite{rogers2010extended}, AlvaDesc descriptors \cite{mauri2020alvadesc} and 3D pharmacophore fingerprints as features and select the best representation for the evaluation. For in-context predictions, we use the same feature sets along with different variants of TabICLv2 \cite{qu2025tabicl}, a tabular foundation model that doesn't require any significant hyperparameter tuning. Further, we evaluate three deep learning models: the descriptor foundation model CheMeleon \cite{green2026deep}, the GNN architecture Chemprop \cite{heid2023chemprop} and the contrastive language-assay trained molecular foundation model CLAMP \cite{seidl2023clamp}. All models are finetuned and hyperparameter tuned using a 5-fold cross-validation on the train split, across the search spaces defined in \ref{a:implementation}. For CLAMP, we use the learned representations directly as features and evaluate both linear and non-linear predictors.

\subsection{Mol-JEPA downstream prediction}
There exist multiple strategies for using Mol-JEPA embeddings for downstream tasks. First, we use the CLS token to extract a global summary of all modalities. We feed this representation of size 512 into a linear and non-linear probe and experiment with TabICLv2 as a predictor.  We use default parameters and do not perform any hyperparameter optimization. Secondly, we use the modality-specific predicted embeddings in an aggregative probe. We choose a 2 layer transformer encoder that receives tokens of size 512 for each modality and applies attention to combine the information. The outputs embeddings are mean pooled and fed into a simple linear output layer. Lastly, we perform LoRA fine-tuning \cite{hu2022lora} with rank 16 and target layers of the transformer module and output projections of the Mol-JEPA architecture. Our experiments, model checkpoints, and Hugging Face integration are publicly available at \href{https://github.com/Boehringer-Ingelheim/mol-jepa}{github.com/Boehringer-Ingelheim/mol-jepa}.

\section{Results}

\begin{figure}[t]
    \centering
    \begin{subfigure}[b]{0.48\linewidth}
        \centering
        \includegraphics[width=\linewidth]{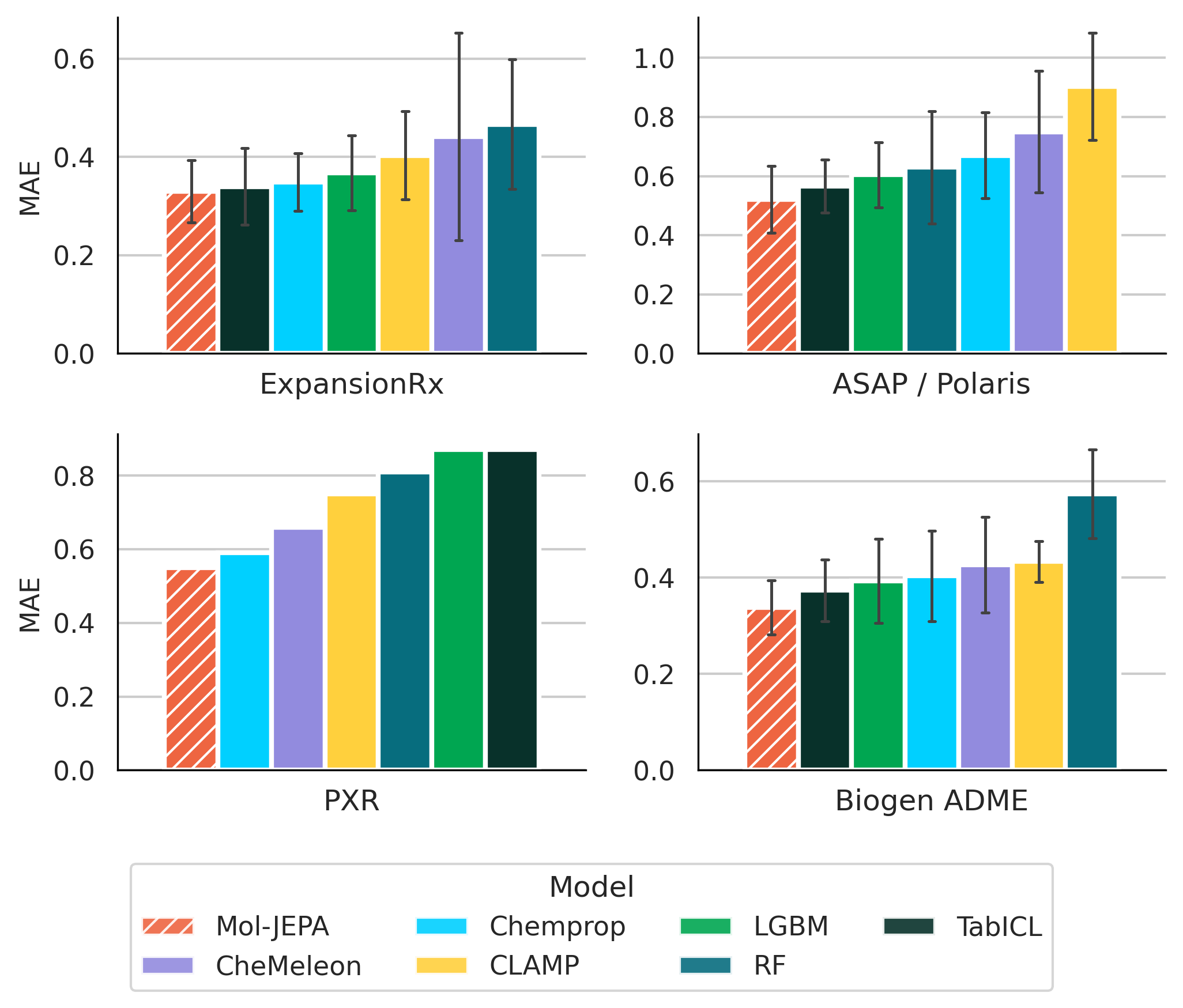}
        \caption{\textbf{Mol-JEPA demonstrates strong performance across tasks.} The bars represent the aggregated mean absolute error of models within each benchmark family along with the standard deviation across the individual datasets.}
        \label{resultsa}
    \end{subfigure}
    \hfill
    \begin{subfigure}[b]{0.48\linewidth}
        \centering
        \includegraphics[width=\linewidth]{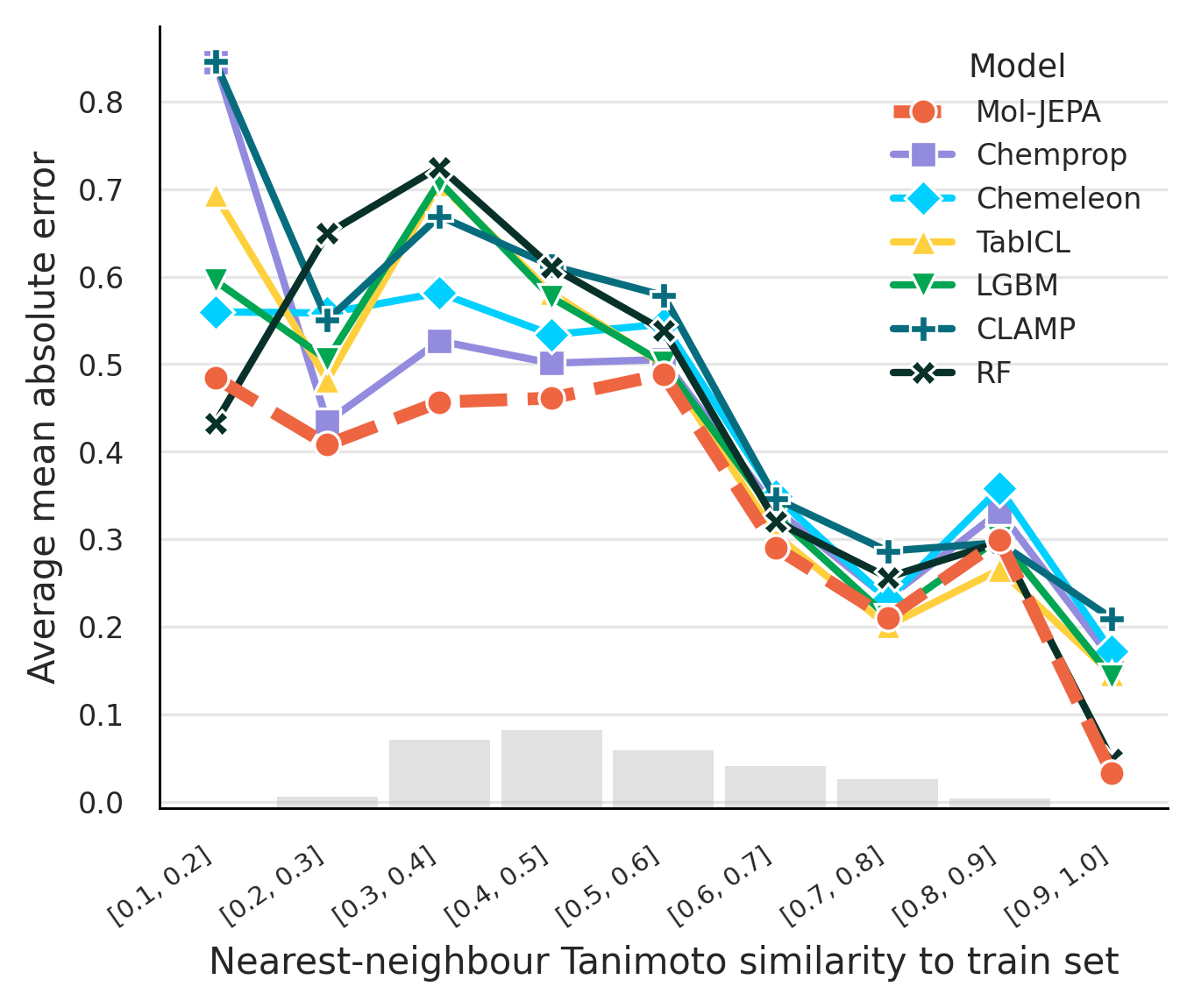}
        \caption{\textbf{Mol-JEPA generalizes better.} The lines show the average MAE for molecules across all datasets, binned by Tanimoto similarity to the train dataset. Mol-JEPA shows the best out-of-distribution performance.}
        \label{resultsb}
    \end{subfigure} 
    \hfill
    \caption{Mol-JEPA evaluation results.}
    \label{fig:results}
\end{figure}

\paragraph{Mol-JEPA outperforms the baselines on multiple datasets.}
As summarized by Figure \ref{resultsa}, Mol-JEPA demonstrates strong performance across all benchmark families. Specifically, our model demonstrates a clear advantage on benchmarks with smaller datasets, such as ASAP/Polaris and Biogen ADME. This finding is especially important in drug discovery, where obtaining large, high-quality labeled datasets is often prohibitively expensive. In addition, we evaluate out-of-distribution performance in Figure \ref{resultsb} and find that Mol-JEPA consistently outperforms the baselines as the distance from the training data increases. This advantage likely arises from pretraining on a large and diverse set of drug-like compounds, allowing Mol-JEPA to capture broad chemical space relevant to drug discovery. A detailed comparison for mean absolute errors is provided in Table \ref{tab:benchmarks} and for additional metrics in Appendix \ref{a:results}.

\begin{figure}[t]
    \centering
    \begin{subfigure}[b]{0.47\linewidth}
        \centering
        \includegraphics[width=\linewidth]{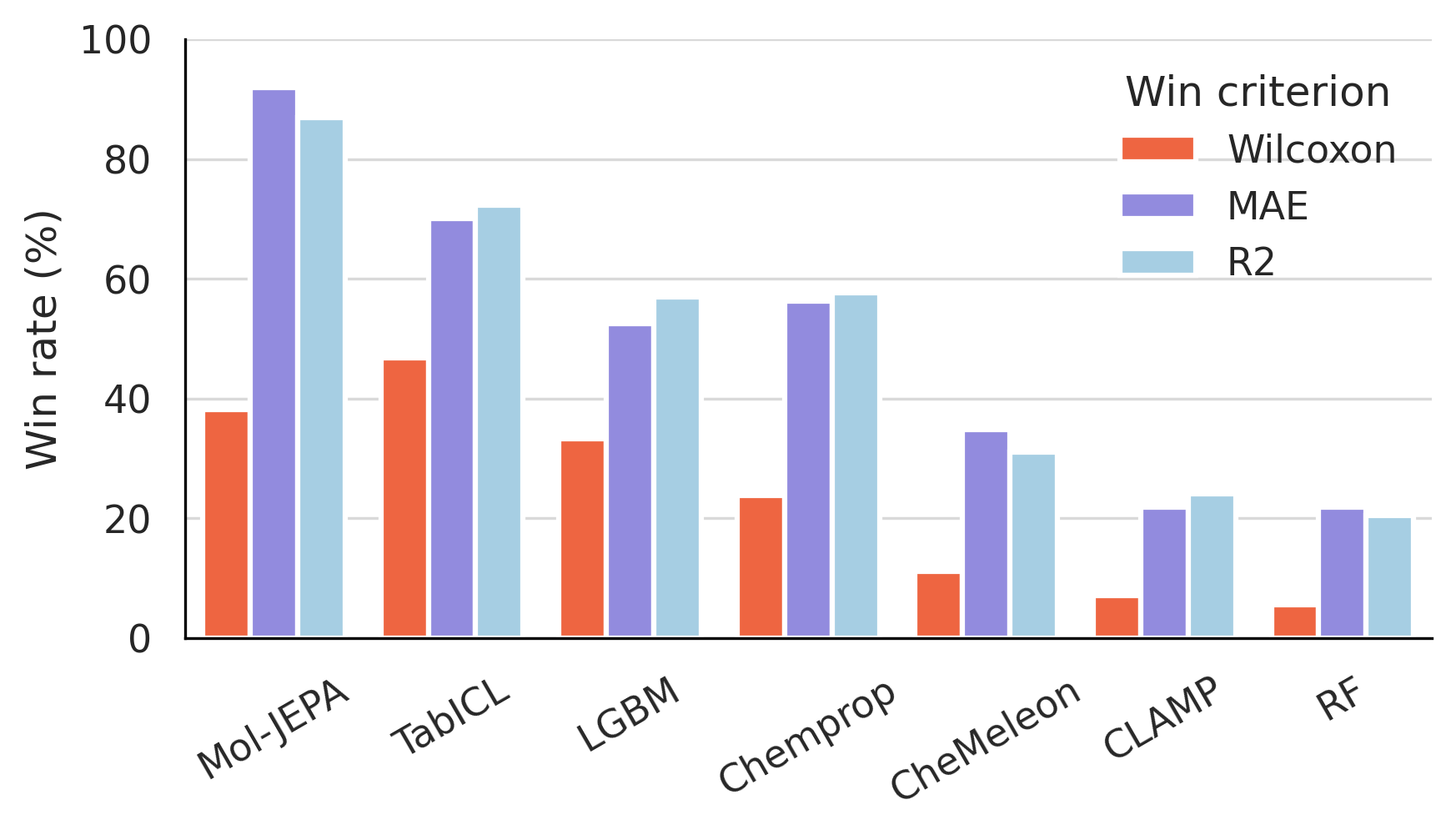}
        \caption{\textbf{Comparison win rates}. The bars show how often a model wins across all pair-wise comparisons. While Mol-JEPA is best using MAE and $R^2$, it wins slightly less often when using Wilcoxon test for evaluation.}
        \label{results2a}
    \end{subfigure}
    \hspace{0.5cm}
    \begin{subfigure}[b]{0.42\linewidth}
        \centering
        \includegraphics[width=\linewidth]{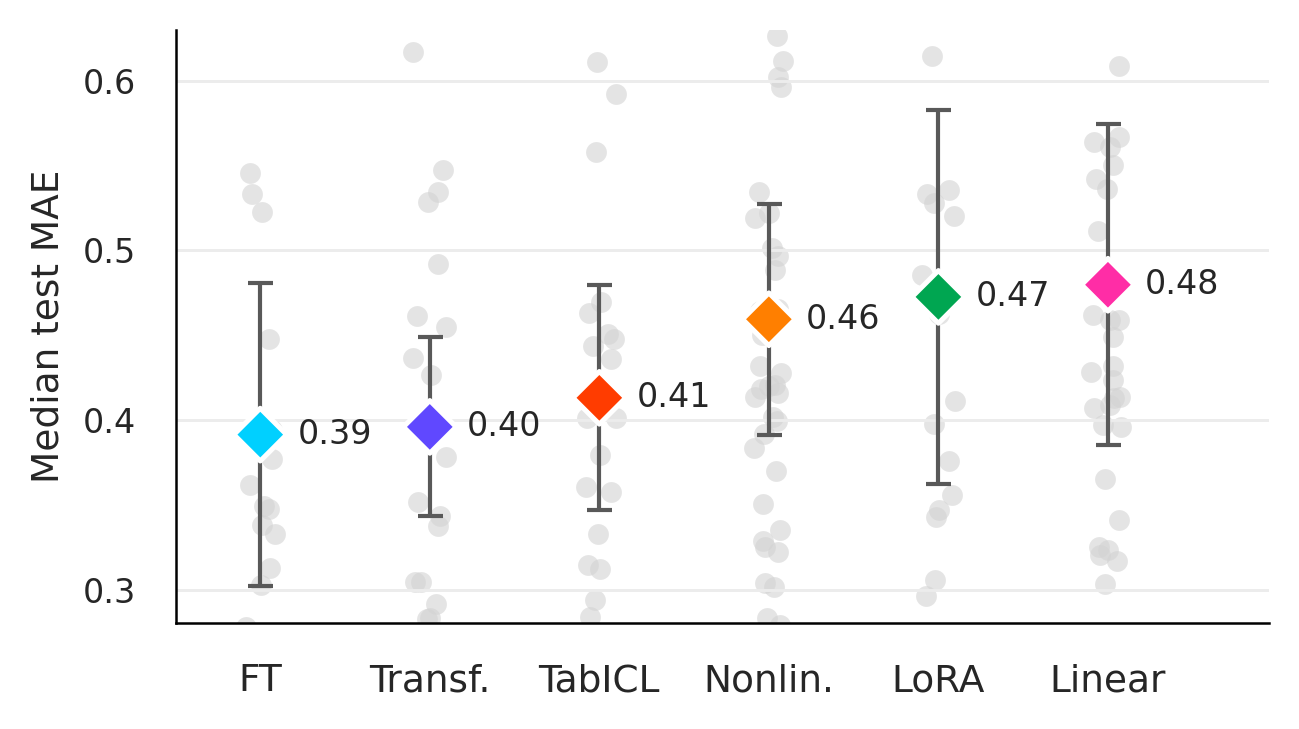}
        \caption{\textbf{Downstream probe performance.} Besides full fine-tuning (FT), the Transformer probe on multimodal embeddings and TabICLv2 on CLS tokens are the best downstream predictors.}
        \label{results2b}
    \end{subfigure} 
    
    \hfill
    \caption{Mol-JEPA training results.}
    \label{fig:results2}
\end{figure}

\paragraph{Statistical significance.}
To assess whether performance differences are statistically significant, we compare the sample-wise absolute errors of all methods using paired Wilcoxon signed-rank tests \cite{wilcoxon1945}. We perform pairwise comparisons between Mol-JEPA and all baseline models and compute mean MAE differences and FDR-corrected p-values. We count the statistically significant pairwise victories across datasets and summarize the win rates in Figure \ref{results2a}, along with the win rates based on MAE and $R^2$ results. We find that Mol-JEPA has the highest win rates when using classical metrics and is close to the highest win rate when using wilcoxon rank tests. Specifically, it achieves superior performance in 38\% of the comparisons, compared to 47\% for TabICL with AlvaDesc descriptors. Further analysis indicates that TabICLv2 derives much of its advantage from the larger ExpansionRx datasets, while Mol-JEPA tends to achieve more significant wins on the smaller ASAP and Biogen benchmarks. These results suggest that Mol-JEPA may offer particular benefits in low-data settings.

\begin{table*}[t]
\centering
\caption{Mean absolute error (MAE) comparison across benchmark datasets for three cluster splits. Mol-JEPA (Best) denotes the best performance achieved across all probe and embedding variants, while Mol-JEPA (Transformer) uses a Transformer head with modality-specific embeddings.}
\label{tab:benchmarks}
\scriptsize
\setlength{\tabcolsep}{4pt}
\renewcommand{\arraystretch}{1.15}

\begin{tabular}{l
>{\columncolor{moljepa}}c
>{\columncolor{moljepa}}c
cccccc}
\toprule
\textbf{Dataset} &
\makecell[c]{\textbf{Mol-JEPA}\\Best} &
\makecell[c]{\textbf{Mol-JEPA}\\Transformer} &
\makecell[c]{\textbf{CLAMP}\\Nonlinear} &
\makecell[c]{\textbf{CheMeleon}\\Finetuned} &
\makecell[c]{\textbf{Chemprop}\\Finetuned} &
\makecell[c]{\textbf{TabICLv2}\\AlvaDesc} &
\makecell[c]{\textbf{RF}\\ECFP4} &
\makecell[c]{\textbf{LGBM}\\AlvaDesc} \\
\midrule \rowcolor{slategray} \multicolumn{9}{l}{\textbf{ExpansionRx}}\\ Caco-2 Permeability & \textbf{0.34}\std{.02} & \underline{0.35}\std{.02} & 0.40\std{.05} & 0.37\std{.05} & 0.39\std{.04} & 0.42\std{.02} & 0.43\std{.02} & 0.42\std{.02} \\ Caco-2 Efflux & \underline{0.25}\std{.02} & 0.26\std{.01} & 0.31\std{.03} & 0.27\std{.05} & \textbf{0.24}\std{.03} & \textbf{0.24}\std{.02} & 0.28\std{.04} & \textbf{0.24}\std{.02} \\ LogD & \textbf{0.33}\std{.02} & 0.46\std{.04} & 0.50\std{.05} & \underline{0.35}\std{.02} & 0.74\std{.06} & 0.36\std{.02} & 0.73\std{.07} & 0.43\std{.04} \\ KSOL & \textbf{0.45}\std{.03} & 0.49\std{.04} & 0.53\std{.04} & \textbf{0.45}\std{.06} & 0.60\std{.05} & \underline{0.46}\std{.06} & 0.62\std{.04} & 0.49\std{.03} \\ HLM CLint & \textbf{0.39}\std{.02} & 0.43\std{.03} & 0.50\std{.02} & 0.40\std{.06} & 0.46\std{.01} & \textbf{0.39}\std{.05} & 0.47\std{.02} & \textbf{0.39}\std{.03} \\ MLM CLint & 0.40\std{.05} & \underline{0.38}\std{.07} & \textbf{0.36}\std{.04} & 0.39\std{.06} & 0.42\std{.04} & 0.41\std{.09} & 0.44\std{.08} & \underline{0.38}\std{.07} \\ MBPB & \underline{0.30}\std{.02} & \underline{0.30}\std{.03} & 0.31\std{.01} & \underline{0.29}\std{.06} & 0.45\std{.10} & \textbf{0.26}\std{.03} & 0.43\std{.01} & 0.34\std{.04} \\ MGMB & \underline{0.30}\std{.05} & \textbf{0.28}\std{.06} & 0.35\std{.08} & 0.31\std{.13} & 0.41\std{.13} & \underline{0.30}\std{.09} & 0.39\std{.12} & 0.33\std{.10} \\ MPPB & \textbf{0.26}\std{.04} & \underline{0.28}\std{.05} & 0.31\std{.03} & 0.31\std{.02} & 0.41\std{.01} & \textbf{0.26}\std{.05} & 0.41\std{.06} & 0.29\std{.05} \\ \midrule \rowcolor{slategray} \multicolumn{9}{l}{\textbf{ASAP}}\\ MERS-CoV-2 Potency & 0.59\std{.10} & 0.70\std{.08} & 0.96\std{.04} & 0.86\std{.09} & \textbf{0.52}\std{.08} & \underline{0.54}\std{.09} & 0.58\std{.04} & 0.59\std{.13} \\ SARS-CoV-2 Potency & \textbf{0.62}\std{.06} & \textbf{0.62}\std{.06} & 1.02\std{.12} & 0.71\std{.13} & 0.73\std{.18} & \underline{0.64}\std{.17} & 0.77\std{.25} & 0.68\std{.16} \\ LogD & \underline{0.68}\std{.09} & 0.72\std{.11} & 1.09\std{.14} & 0.84\std{.19} & \textbf{0.64}\std{.14} & 0.89\std{.07} & 0.99\std{.05} & 0.74\std{.14} \\ KSOL & \textbf{0.42}\std{.19} & \underline{0.53}\std{.08} & 0.91\std{.19} & 0.60\std{.02} & 0.67\std{.19} & 0.65\std{.12} & 0.58\std{.06} & 0.70\std{.11} \\ HLM & \underline{0.39}\std{.11} & \underline{0.39}\std{.11} & 0.58\std{.03} & 0.50\std{.24} & \textbf{0.36}\std{.09} & 0.42\std{.15} & 0.43\std{.07} & 0.43\std{.10} \\ MLM & 0.53\std{.01} & 0.53\std{.01} & 1.02\std{.37} & 0.67\std{.13} & 0.64\std{.08} & 0.59\std{.15} & \textbf{0.50}\std{.06} & \underline{0.51}\std{.09} \\ MDR1 Efflux & \textbf{0.42}\std{.13} & \underline{0.44}\std{.14} & 0.74\std{.14} & 0.51\std{.30} & 0.45\std{.22} & 0.48\std{.26} & 0.56\std{.36} & 0.58\std{.41} \\ \midrule \rowcolor{slategray} \multicolumn{9}{l}{\textbf{PXR}}\\ PXR Activity & \textbf{0.55}\std{.04} & \textbf{0.55}\std{.06} & 0.75\std{.06} & \underline{0.59}\std{.13} & 0.79\std{.11} & 0.87\std{.20} & 0.81\std{.11} & 0.87\std{.17} \\ \midrule \rowcolor{slategray} \multicolumn{9}{l}{\textbf{Biogen ADME}}\\ Solubility & \textbf{0.30}\std{.02} & \textbf{0.30}\std{.02} & 0.38\std{.02} & 0.35\std{.03} & 0.42\std{.04} & \underline{0.33}\std{.01} & 0.43\std{.01} & \underline{0.33}\std{.01} \\ HLM CLint & \textbf{0.33}\std{.02} & \underline{0.34}\std{.02} & 0.45\std{.06} & 0.35\std{.01} & 0.51\std{.09} & 0.37\std{.03} & 0.55\std{.08} & 0.36\std{.02} \\ RLM CLint & \textbf{0.37}\std{.04} & \underline{0.38}\std{.04} & 0.46\std{.04} & 0.39\std{.01} & 0.57\std{.07} & 0.39\std{.03} & 0.56\std{.04} & 0.39\std{.04} \\ HPPB & \textbf{0.33}\std{.06} & \underline{0.34}\std{.07} & 0.45\std{.05} & 0.48\std{.08} & 0.56\std{.07} & 0.35\std{.04} & 0.67\std{.11} & 0.40\std{.07} \\ RPPB & \textbf{0.43}\std{.07} & \underline{0.45}\std{.08} & 0.48\std{.18} & 0.55\std{.13} & 0.57\std{.18} & 0.49\std{.12} & 0.68\std{.15} & 0.56\std{.12} \\ MDR1 Efflux & \textbf{0.27}\std{.03} & \underline{0.29}\std{.01} & 0.38\std{.05} & 0.30\std{.06} & 0.39\std{.02} & 0.31\std{.04} & 0.55\std{.08} & 0.32\std{.04} \\ \midrule \rowcolor{slategray} \textbf{Average MAE} & \textbf{0.402} & \underline{0.427} & 0.576 & 0.471 & 0.519 & 0.453 & 0.559 & 0.468 \\
\bottomrule
\end{tabular}
\end{table*}

\paragraph{Downstream probe evaluation.}
Figure \ref{results2b} summarizes the performance of Mol-JEPA embeddings using different downstream strategies. While full fine-tuning has the best median error, we find that a lightweight multimodal Transformer and TabICLv2 on CLS tokens are the strongest downstream models. More light-weight models, such as a linear or non-linear probes generally lead to worse performance, which suggests that the representations are too complex for these models. Finally, LoRA-finetuning of the Mol-JEPA transformer module under-performs most other methods. This is presumably due to hyperparameter sensitivity or overfitting.

\begin{figure}[t]
    \centering
    \begin{subfigure}[b]{0.45\linewidth}
        \centering
        \includegraphics[width=\linewidth]{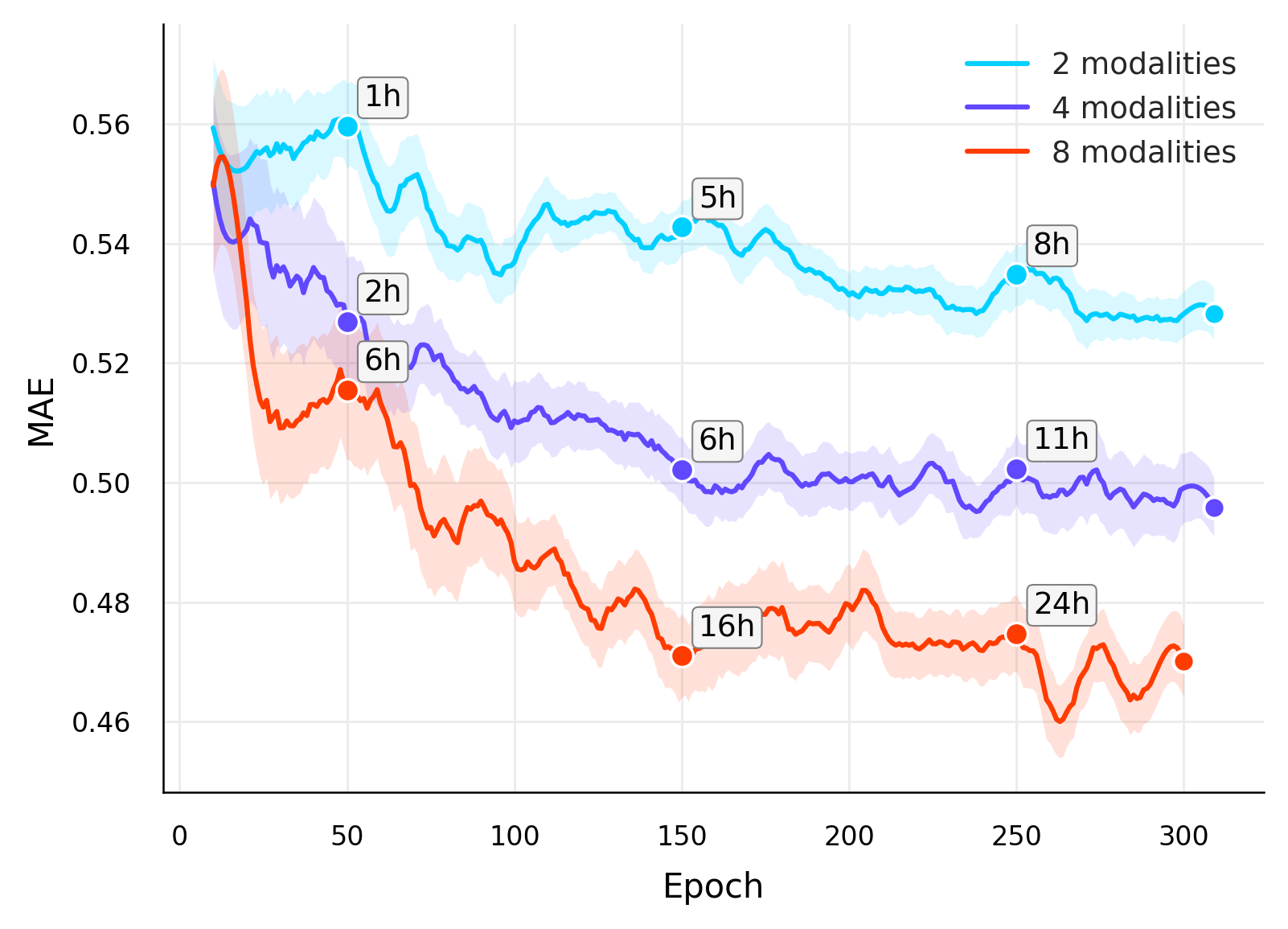}
        \caption{\textbf{Performance increases with more modalities.} The lines represent the training time and epochs on a 100k subset dataset, annotated with the training hours.}
        \label{fig:modalitiesa}
    \end{subfigure}
    \hspace{0.5cm}
    \begin{subfigure}[b]{0.45\linewidth}
        \centering
        \includegraphics[width=\linewidth]{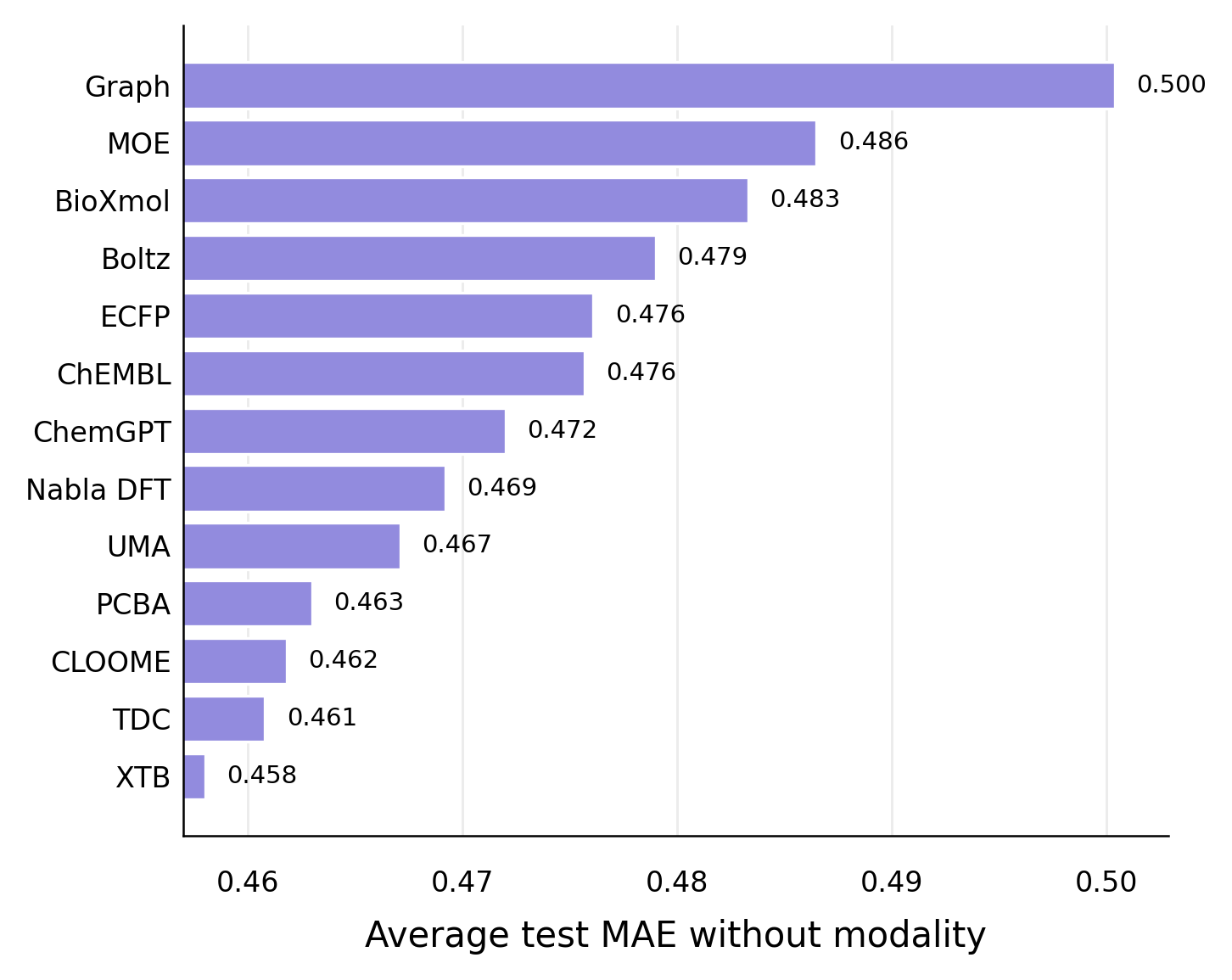}
        \caption{\textbf{Leave-one-modality-out analysis.} We analyze the effect of removing modalities during training. The reported scores are the average downstream probe MAEs.}
        \label{fig:modalitiesb}
    \end{subfigure} 
    \hfill
    \caption{Modality analysis}
    \label{fig:modalities}
\end{figure}

\paragraph{Modality scaling in Mol-JEPA} As shown in Figure \ref{fig:modalitiesa}, training time increases approximately linearly with the number of modalities, although it also depends on the specific modalities used and their input dimensionalities. For instance, when pretraining on a dataset of 100k molecules for 300 epochs on a single GPU, using two modalities (CLOOME and Graph) requires roughly 9 hours, four modalities (adding UMA and MOE) require about 13 hours, and eight modalities (further adding ECFP, Boltz, BioXMol, and ChemGPT) increase the training time to approximately 27 hours. In all experiments, we observe a better performance when increasing the number of modalities, as annotated in Figure \ref{fig:modalitiesa} and Appendix Figure \ref{modalitygain}. These results highlight the effectiveness of Mol-JEPA as a multimodal learning framework, showing that the integration of additional modalities consistently enhances downstream task performance.

\paragraph{Leave-one-modality-out analysis}
To better understand the contribution of individual modalities, we perform a leave-one-modality-out analysis. We observe only minor performance differences when removing a single modality at inference time, suggesting that the learned representations contain a certain degree of redundancy. This behavior is likely encouraged by the reconstruction objective, which promotes the encoding of overlapping information across modalities. To better quantify the contribution of individual modalities, we perform a leave-one-modality-out analysis during training on a 100k subset of the pretraining data. In each experiment, one modality is completely removed, allowing us to measure the impact of its full absence on representation learning. Figure~\ref{fig:modalities} summarizes the resulting downstream errors, where larger error increases indicate a greater contribution of the removed modality to downstream performance. We find that the graph modality contributes the most to downstream performance. One possible explanation is that it is the only modality with a trainable GNN backbone, whereas the remaining modalities rely on frozen representations. We also find that the molecular fingerprint modalities MOE and ECFP4 rank among the most important modalities, indicating that their expert-designed features contain valuable information for downstream tasks. Lastly, we find that modalities capturing chemical knowledge through binding affinities and cellular interaction profiles also contribute to the learned representations, highlighting the value of incorporating biological context alongside molecular structure. We discuss these results and further experiments in Appendix \ref{a:ablations}.

\begin{figure}[t]
    \centering
    \includegraphics[width=0.95\linewidth]{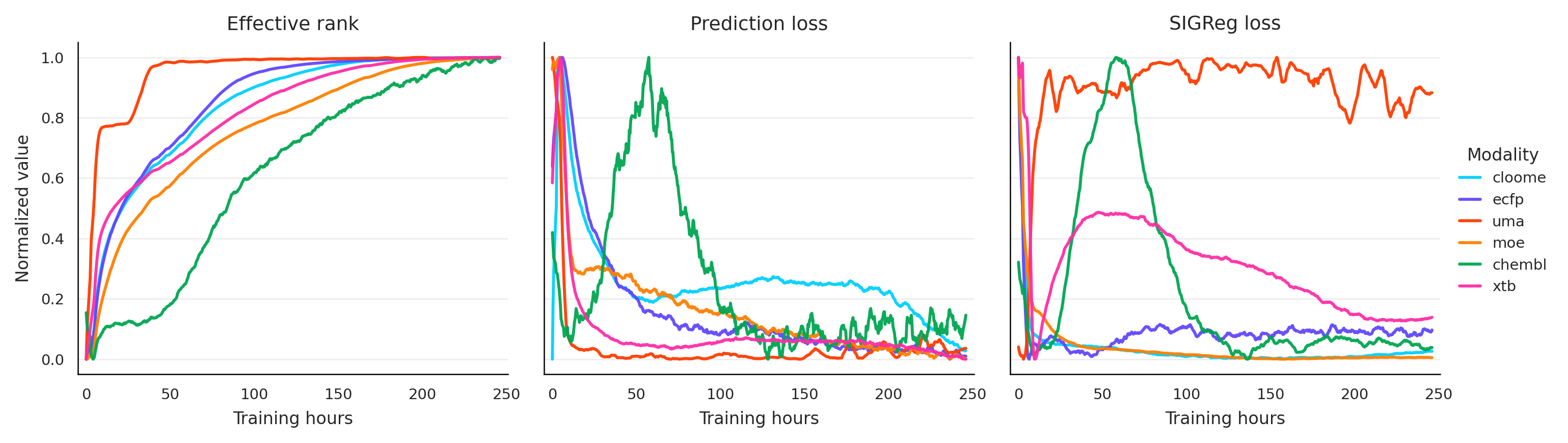}
    \caption{\textbf{Multimodal training dynamics.} Effective rank, prediction loss and SIGReg loss for different modalities during training. While effective rank is continuously increasing, the losses exhibit instabilities for some modalities.}
    \label{fig:metrics}
\end{figure}

\begin{figure}[ht]
    \centering
    \includegraphics[width=0.4\linewidth]{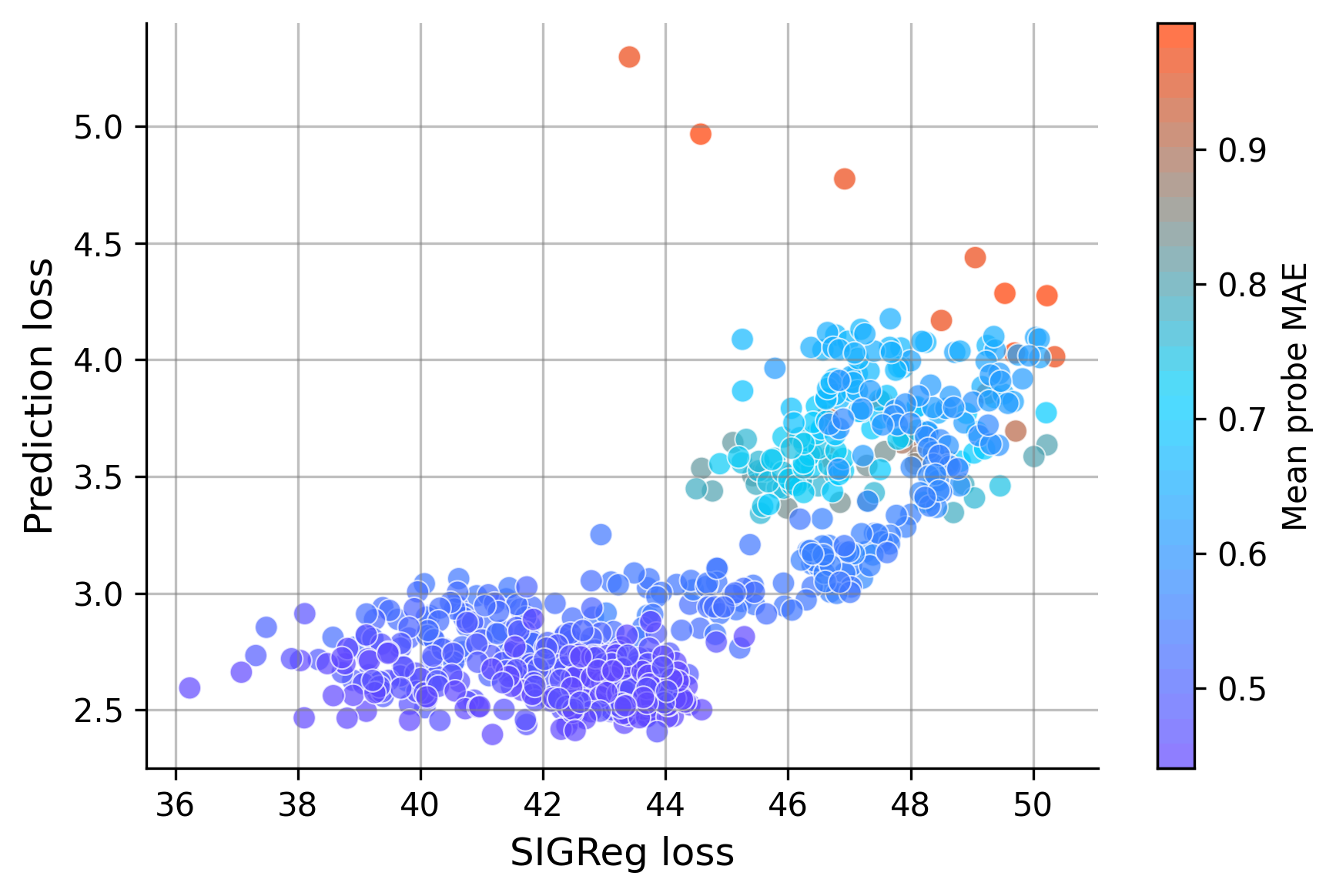}
    \caption{Losses and downstream error.}
    \label{fig:losses}
\end{figure}

\paragraph{Multimodal training dynamics.}
Our proposed architecture enables stable multimodal training, and we observe a consistent relationship between the prediction loss, the SIGReg loss, and downstream performance, as illustrated in Figure \ref{fig:losses}. However, as shown in Figure \ref{fig:metrics}, the training dynamics differ across modalities. While the effective rank increases consistently for all modality-specific embeddings, the prediction loss and isotropic regularization exhibit more heterogeneous behavior, suggesting that modalities contribute differently to the learning process. For instance, the experimental ChEMBL vectors first experience a rise in the prediction error and then improve steadily throughout training. In contrast, UMA embeddings show a decreasing prediction loss but oscillating regularization loss. We also find that sparse modalities are generally more difficult to optimize as their loss signal is less dominant compared to modalities that are always available. Lastly, we find that the inclusion of specific modalities, such as CLOOME can decrease the downstream performance. The experienced complexity of multimodal pretraining is in line with recent works \cite{kamai2026align}, showing that different modalities might require different approaches and not all might be suitable for joint embedding predictive architectures.

\section{Conclusion}
\label{sec:conclusion}
In this work, we presented \textbf{Mol-JEPA}, a multimodal foundation model that enables stable learning from many molecular modalities by predicting their latent representations. Using various high quality benchmarks, we show that Mol-JEPA achieves strong performance across different downstream tasks. Furthermore, we statistically compare the predictions and find that our model is particularly better on smaller datasets and in out-of-distribution settings. Additionally, we analyze the impact of individual modalities on training dynamics and downstream performance, suggesting that the selection of the most effective modality set remains an important direction for future research. Given the limited data availability in drug discovery, we believe this work represents an important step toward molecular world models that combine diverse sources of information to reason about physical dynamics and biochemical interactions. More broadly, Mol-JEPA demonstrates a general approach to multimodal representation learning that could be applied across scientific domains, including materials science, biology, and other data-scarce fields.

\section{Limitations}

There are several limitations and opportunities for future research. First, the current reconstruction objective does not account for the fact that some modalities may be inherently unable to recover information contained in others. Incorporating modality-specific dependencies and advanced masking strategies into the training objective may therefore be beneficial. Second, as with most machine learning models, out-of-distribution generalization remains a key challenge. Future work should investigate the robustness of Mol-JEPA when applied to molecules, tasks, and data distributions that differ substantially from those encountered during pretraining. Finally, our understanding of multimodal learning dynamics during training and inference remains limited. In particular, it is still unclear how individual modalities contribute to downstream performance and how factors such as data sparsity, modality dimensionality, information content, and redundancy affect representation learning. Addressing these questions could lead to more effective modality selection and improved multimodal training strategies. Lastly, due to computational constraints, most ablation studies were conducted on a considerably smaller dataset than the full pretraining data, potentially impacting some conclusions.

\bibliography{iclr2026_conference}
\bibliographystyle{iclr2026_conference}

\appendix

\renewcommand{\appendixname}{Appendix}

\newpage
\section*{Appendices}
\addcontentsline{toc}{section}{Appendices} 
\startcontents[appendices]
\printcontents[appendices]{l}{1}{\setcounter{tocdepth}{2}} 
\vspace{1cm}
\newpage

\section{Dataset details}
\label{a:datasets}

\paragraph{ChEMBL processing}

\begin{wraptable}{r}{0.45\textwidth}
  \centering
  \caption{Assay threshold filtering.}
  \label{tab:unique_targets}
  \begin{tabular}{lr}
    \toprule 
    Threshold & Remaining assays \\ 
    \midrule
    1    & 37,178 \\ 
    10   & 904    \\ 
    20   & 349    \\ 
    50   & 103    \\ 
    100  & 34     \\ 
    500  & 17     \\ 
    1000 & 24     \\ 
    \bottomrule 
  \end{tabular}
\end{wraptable}

We construct a feature matrix by pivoting ChEMBL bioassay activity data, representing each molecule as a vector of assay measurements. To ensure sufficient coverage of individual assay endpoints while maintaining a diverse set of biological readouts, we analyze the trade-off between assay coverage and data sparsity. As shown in Table \ref{tab:unique_targets}, increasing the minimum number of required measurements per assay reduces the number of retained endpoints. Based on this analysis, we select a threshold of 30 measurements per assay, resulting in a feature vector with 306 dimensions. The retained assays encompass a broad range of ADMET and physicochemical properties, including microsomal stability, solubility, hERG channel inhibition, receptor binding activities, and other pharmacologically relevant endpoints across multiple species.

\paragraph{Therapeutics Data Commons processing}
Table \ref{tab:tdc_summary} provides an overview of all processed datasets. Analogous to the ChEMBL data, measurements are aggregated at the molecule level, resulting in a feature vector of all available experimental observations. Since several datasets contain multiple endpoints, the resulting feature vectors have 672 dimensions. The number of available measurements varies substantially across datasets, ranging from as few as 50 observations for certain toxicity endpoints to more than 300,000 measurements in high-throughput screening (HTS) datasets.

\begin{table}[ht]
\centering
\caption{TDC dataset summary}
\label{tab:tdc_summary}
\small
\begin{tabular}{p{0.29\textwidth} p{0.52\textwidth} c}
\hline
\textbf{Dataset} & \textbf{Description} & \textbf{Category} \\
\hline
Caco-2 & Cell permeability measure & ADME \\
PAMPA & Parallel artificial membrane permeability & ADME \\
HIA & Human intestinal absorption classification & ADME \\
P-gp & P-glycoprotein inhibition activity & ADME \\
Bioavailability & Rate and extent of drug absorption & ADME \\
Lipophilicity & Lipid solubility measure & ADME \\
Solubility (AqSolDB) & Aqueous solubility in water & ADME \\
Hydration Free Energy & Solvation energy in water & ADME \\
BBB Penetration & Blood-Brain Barrier passage & ADME \\
PPBR & Plasma protein binding percentage & ADME \\
VDss & Volume of distribution at steady state & ADME \\
CYP Inhibition & Cytochrome P450 inhibition (1A2, 2C9, 2C19, 2D6, 3A4) & ADME \\
CYP Substrates & Cytochrome P450 substrate activity (2C9, 2D6, 3A4) & ADME \\
Half-Life & Elimination half-life in body & ADME \\
Hepatocyte Clearance & Hepatic metabolic clearance rate & ADME \\
\hline
LD50 & Median lethal dose (acute toxicity) & Tox \\
hERG Cardiotoxicity & Potassium channel blockade \& cardiac risk & Tox \\
Ames Test & Bacterial mutagenicity assay & Tox \\
DILI & Drug-induced liver injury risk & Tox \\
Skin Reaction & Dermal allergic response & Tox \\
Carcinogenicity & Cancer-promoting potential & Tox \\
Tox21 / ToxCast & High-throughput nuclear \& pathway toxicity & Tox \\
ClinTox & Clinical trial failures due to toxicity & Tox \\
\hline
SARS-CoV-2 & In-vitro viral inhibition \& 3CLPro protease screen & HTS \\
HIV & HIV replication inhibition & HTS \\
Orexin-1 Receptor & Target binding \& modulation & HTS \\
M1 Muscarinic & Receptor agonist \& antagonist activity & HTS \\
Ion Channels & Potassium (Kir2.1, KCNQ2) and Calcium (Cav3) channels & HTS \\
Choline Transporter & Neurotransporter binding & HTS \\
STK33 \& TDP1 & Kinase and phosphodiesterase enzymatic screens & HTS \\
\hline
\end{tabular}
\end{table}

\paragraph{Benchmark data processing}
All benchmark datasets are downloaded and used in the form provided by the challenge organizers, as the majority of the required preprocessing has already been performed. For ExpansionRx, it was additionally necessary to convert some endpoints to a logarithmic scale by applying $\log_{10}(x + 0.001)$, consistent with the preprocessing used in the competition, where an epsilon of $0.001$ was added prior to the transformation.

\begin{figure}[t]
    \centering
    \includegraphics[width=0.95\linewidth]{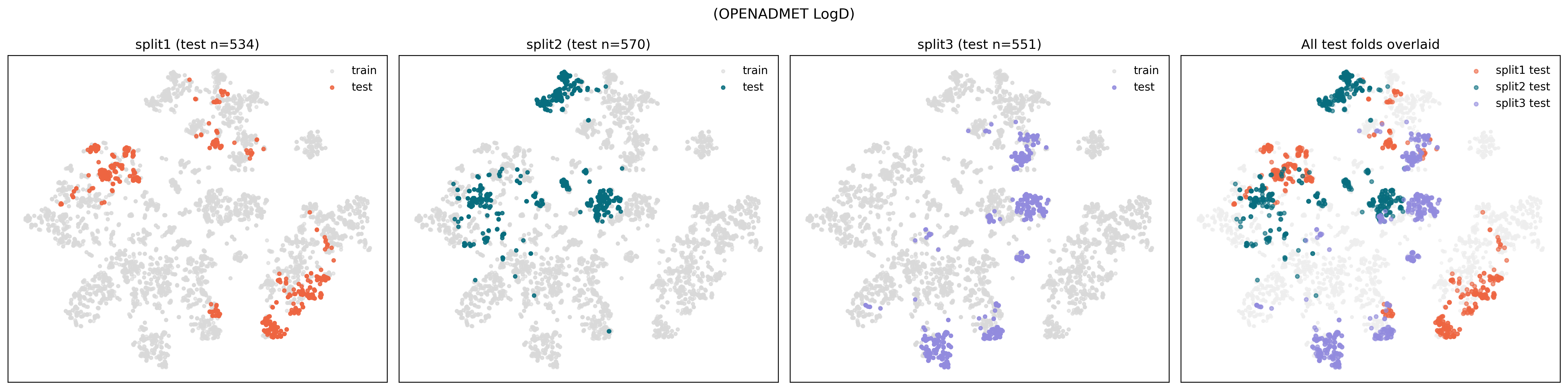}
    \caption{Cluster splits t-SNE visualization for ExpansionRx LogD dataset using ECFP4 fingerprints.}
    \label{fig:splits}
\end{figure}

\paragraph{Dataset splitting}
In Table \ref{tab:clustersplits} we provide the number of samples for each benchmark dataset along with the dataset sizes for each split. We move specific clusters to the test data and keep the remaining ones in the training data. In Figure \ref{fig:splits}, we visualize the three splits on the OpenADMET LogD dataset. Although Butina clustering at a Tanimoto similarity threshold of 0.65 reduces structural overlap between training and test sets, it does not strictly enforce a maximum cross-split similarity of 0.65 and some molecule pairs assigned to different clusters may still exceed this threshold. To assess the impact of such cases, we additionally applied a strict similarity-based filtering procedure to remove all train-test pairs above the threshold and found that this had no substantial effect on the experimental results.

\begin{table}[t]
\centering
\caption{Train/test sample counts for the three cluster splits of each benchmark task.}
\label{tab:clustersplits}
\small
\begin{tabular}{llrrrrrrr}
\toprule
Dataset & Task & Total & \multicolumn{2}{c}{Split 1} & \multicolumn{2}{c}{Split 2} & \multicolumn{2}{c}{Split 3} \\
\cmidrule(lr){4-5} \cmidrule(lr){6-7} \cmidrule(lr){8-9}
 &  &  & Train & Test & Train & Test & Train & Test \\
\midrule
ASAP ADMET & HLM Clearance & 166 & 148 & 18 & 153 & 13 & 153 & 13 \\
           & Kinetic Solubility & 214 & 183 & 31 & 186 & 28 & 185 & 29 \\
           & LogD & 203 & 171 & 32 & 163 & 40 & 169 & 34 \\
           & MDR1 Efflux & 242 & 210 & 32 & 205 & 37 & 213 & 29 \\
           & MLM Clearance & 189 & 150 & 39 & 157 & 32 & 145 & 44 \\
\midrule
ASAP Potency & MERS M$^{\text{pro}}$ & 421 & 383 & 38 & 366 & 55 & 366 & 55 \\
             & SARS M$^{\text{pro}}$ & 356 & 327 & 29 & 307 & 49 & 335 & 21 \\
\midrule
Biogen ADME & HLM Clearance & 3087 & 2983 & 104 & 3012 & 75 & 3012 & 75 \\
            & MDR1 Efflux & 2642 & 2565 & 77 & 2567 & 75 & 2566 & 76 \\
            & Human PPB & 194 & 169 & 25 & 174 & 20 & 173 & 21 \\
            & Rat PPB & 168 & 148 & 20 & 148 & 20 & 148 & 20 \\
            & RLM Clearance & 3054 & 2947 & 107 & 2930 & 124 & 2950 & 104 \\
            & Solubility & 2173 & 2075 & 98 & 2060 & 113 & 2048 & 125 \\
\midrule
OpenADMET & Caco-2 Efflux Ratio & 2161 & 1822 & 339 & 1901 & 260 & 1953 & 208 \\
          & Caco-2 $P_{\text{app}, A\rightarrow B}$ & 2156 & 1918 & 238 & 1942 & 214 & 1954 & 202 \\
          & HLM Clearance & 3595 & 3152 & 443 & 3186 & 409 & 3169 & 426 \\
          & Kinetic Solubility & 5128 & 4520 & 608 & 4632 & 496 & 4608 & 520 \\
          & LogD & 5039 & 4505 & 534 & 4469 & 570 & 4488 & 551 \\
          & Mouse Brain Unbound Fraction & 973 & 853 & 120 & 860 & 113 & 797 & 176 \\
          & Mouse Gut Homogenate Binding & 221 & 206 & 15 & 199 & 22 & 177 & 44 \\
          & MLM Clearance & 4375 & 3758 & 617 & 3847 & 528 & 3720 & 655 \\
          & Mouse Plasma Unbound Fraction & 1292 & 1159 & 133 & 1145 & 147 & 1146 & 146 \\
          & PXR ($\text{pEC}_{50}$) & 4288 & 4043 & 245 & 4055 & 233 & 4052 & 236 \\
\bottomrule
\end{tabular}
\end{table}

\section{Implementation details}
\label{a:implementation}

\begin{figure*}[tbp] \centering \begin{minipage}[t]{0.58\textwidth} \centering \vspace{0pt} \scriptsize \renewcommand{\arraystretch}{1.1} \begin{tabular}{l l l l} \toprule  \textbf{Component} & \textbf{Hyperparameter} & \textbf{Search Space} & \textbf{Best Value} \\ \midrule Optimizer & Learning rate & \{0.01, 0.005, 0.001, 0.0005\} & 0.0005 \\ Optimizer & Weight decay & \{0.0, 0.0001, 0.001, 0.01\} & 0.01 \\ \midrule Mol-JEPA Transformer & Hidden dimension & \{32, 128, 512\} & 512 \\ Mol-JEPA Transformer & Masking ratio & \{0.1, 0.3, 0.6, 0.9\} & 0.3 \\ Mol-JEPA Transformer & $\lambda$ & \{0.1, 0.3, 0.5, 0.7\} & 0.7 \\ Mol-JEPA Transformer & Dropout & \{0.1, 0.4, 0.7\} & 0.1 \\ \midrule Graph Encoder & Number of layers & \{1, 2, 3\} & 3 \\ Graph Encoder & Pooling & \{\texttt{mean}, \texttt{max}\} & \texttt{max} \\ Graph Encoder & Attention heads & \{2, 4, 8\} & 8 \\ \midrule Embedding Encoder & Number of layers & \{1, 2, 3\} & 3 \\ \midrule Atom Encoder & Number of layers & \{1, 2, 3\} & 1 \\ Atom Encoder & Attention heads & \{2, 4, 8\} & 4 \\ \midrule Data Loader & Batch size & \{32, 64, 128\} & 128 \\ \bottomrule \end{tabular} \captionof{table}{Hyperparameter search space used for model tuning.} \label{tab:hparam_search} \end{minipage} \hfill \begin{minipage}[t]{0.38\textwidth} \centering \vspace{0pt} \includegraphics[width=\linewidth]{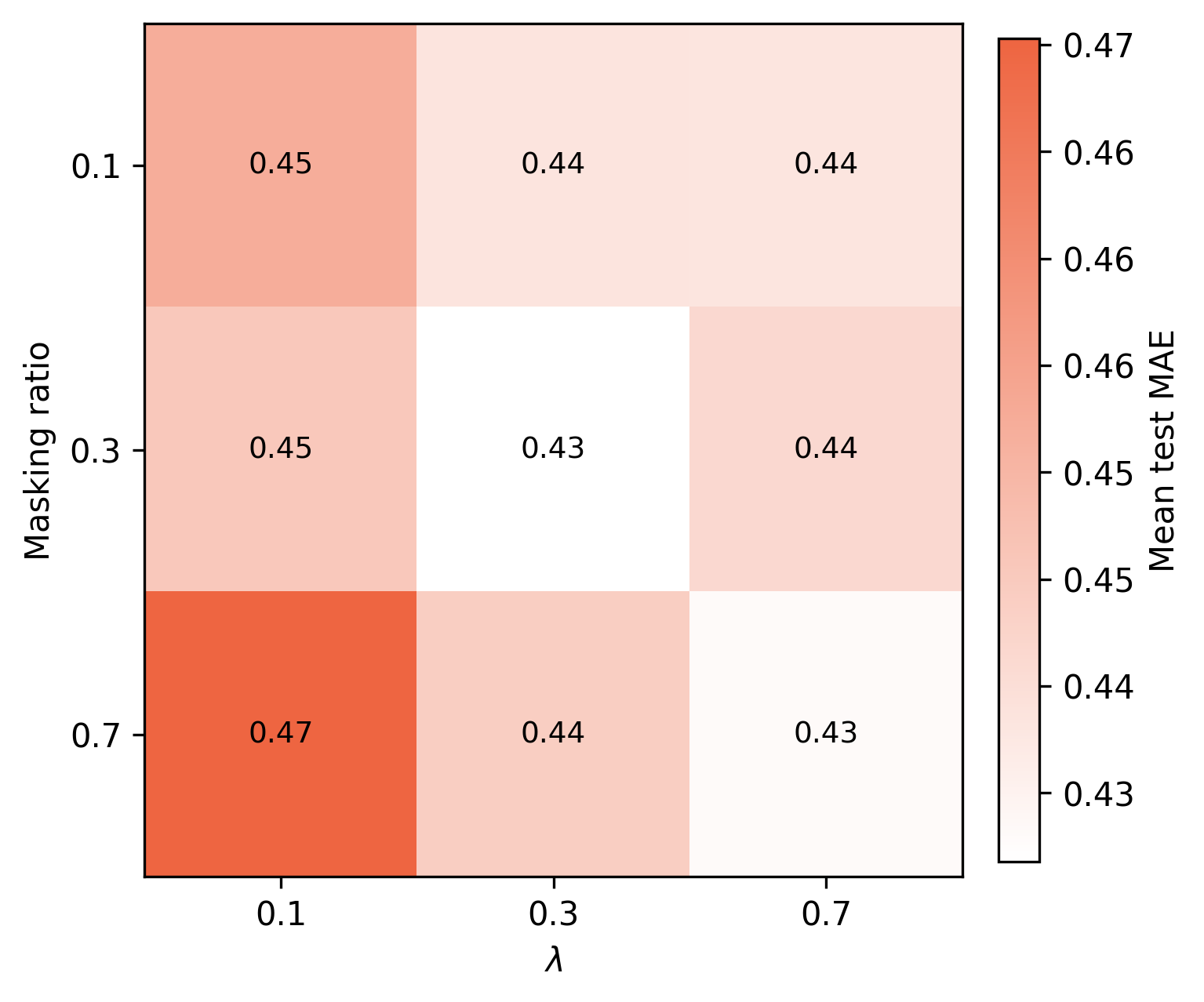} \captionof{figure}{Loss parameters performance.} \label{fig:ablationgrid} \end{minipage} \end{figure*}

\paragraph{Hyperparameter tuning}
Table \ref{tab:hparam_search} summarizes the hyperparameter search space used for tuning Mol-JEPA. Hyperparameter optimization was performed on a subset of 100,000 molecules by evaluating 300 randomly sampled configurations in parallel and selecting the model with the best online probe performance. The most important architectural hyperparameters include the dimensions of the modality-specific encoders, the structure of the Mol-JEPA aggregation module, and standard neural network training parameters. In addition, we explored different masking ratios and values of the loss-balancing parameter $\lambda$. As shown in Figure \ref{fig:ablationgrid}, performance deteriorates for small values of $\lambda$, whereas no consistent relationship between masking ratio and downstream performance could be observed.

\begin{figure}[t]
    \centering
    \includegraphics[width=1\linewidth]{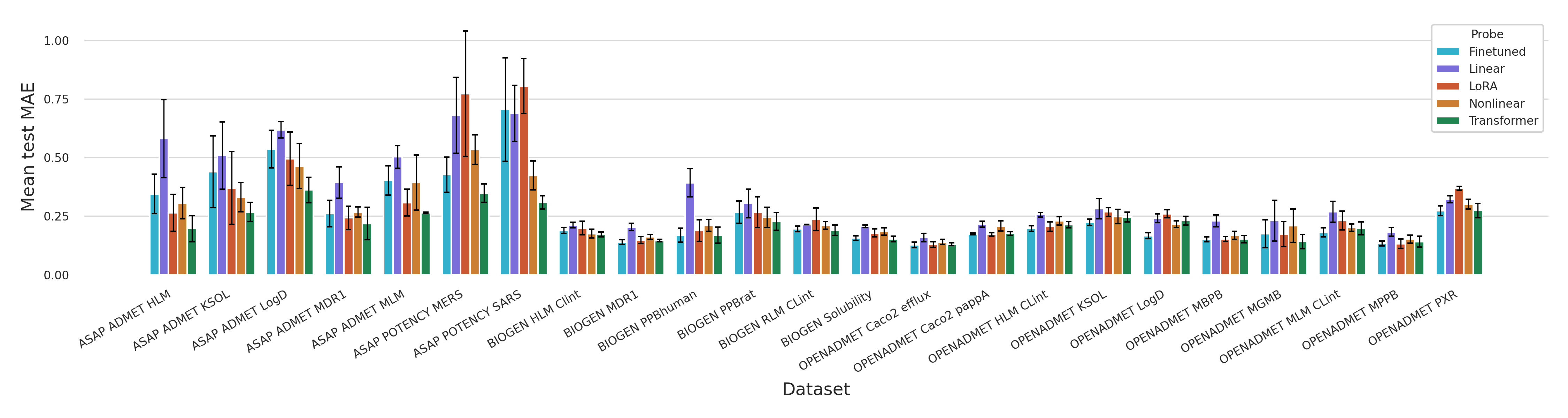}
    \caption{Mean absolute error of different probe variants across datasets.}
    \label{extraction}
\end{figure}

\begin{figure}[t]
    \centering
    \includegraphics[width=1\linewidth]{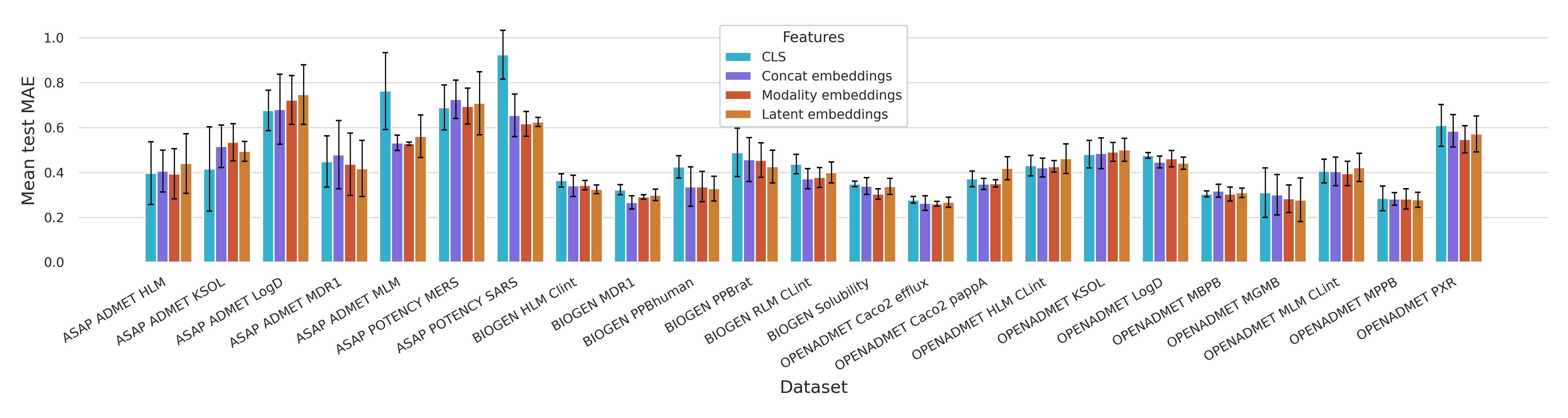}
    \caption{Mean absolute error of different feature variants across datasets.}
    \label{probes}
\end{figure}

\paragraph{Probing model details}
We evaluate several strategies for leveraging Mol-JEPA representations in downstream prediction tasks. The most straightforward approach is to use the CLS token as a global representation of the molecule. This 512-dimensional embedding is provided as input to either linear or nonlinear neural network predictors. Additionally, we employ TabICLv2 as a more expressive downstream model, using $n_{\mathrm{estimators}}=8$. A second strategy utilizes the modality-specific embeddings predicted by Mol-JEPA. Beyond using the final-layer embeddings, we investigate whether intermediate representations contain complementary information by extracting features prior to the final readout layer. We further evaluate a concatenated representation that combines intermediate and final-layer embeddings. As a third approach, we fine-tune the Mol-JEPA transformer module using low-rank adaptation (LoRA). For all fine-tuning experiments, we set the LoRA rank to 16, train for 60 epochs, and use a learning rate of $5 \times 10^{-5}$. Figure~\ref{extraction} compares the performance of the different feature extraction strategies, while Figure~\ref{probes} reports the predictive performance of the downstream models. Overall, the best results are obtained either by combining the CLS token representation with TabICLv2 or by using multimodal embeddings with the transformer-based downstream model.

\paragraph{Computational requirements} 
Mol-JEPA is trained on 12 RTX-6000ADA GPUs with parallelized data loading using 8 CPUs and 128GB of memory. The model is optimized with single precision (float32) and we use 2 gradient accumulation steps along with an aggregated batch size of 1024. The training converges after 3 days at around 150 epochs. After this point, we observe overfitting and degradation of online probe performance.

\section{Baseline models}
\label{a:baselines}
Table \ref{tab:hyperparameter_search_spaces} summarizes the hyperparameter search spaces considered for tuning the baseline models. All feature-based models were trained using ECFP4 fingerprints, AlvaDesc descriptors, and 3D pharmacophore fingerprints as input representations. To reduce the risk of overfitting, the number of AlvaDesc descriptors (nearly 6000) was limited to at most one quarter of the dataset size. Feature selection was performed using Scikit-learn's \texttt{SelectKBest} method with both mutual information and F-test scores, after which the selected descriptor sets were combined. For each dataset split, hyperparameter optimization was conducted using five-fold cross-validation on the training data, and the best-performing model was subsequently evaluated on the corresponding test split.

\begin{table}[ht] \centering \caption{Hyperparameter search space used for baseline model tuning.} \label{tab:hyperparameter_search_spaces} \small \renewcommand{\arraystretch}{1.1} \begin{tabular}{l l l} \toprule \textbf{Model} & \textbf{Hyperparameter} & \textbf{Search Space} \\ \midrule Random Forest & Number of estimators & \{100, 200, 300, 400, 500\} \\ Random Forest & Criterion & \{\texttt{squared\_error}, \texttt{absolute\_error}, \texttt{friedman\_mse}, \texttt{poisson}\} \\ Random Forest & Maximum depth & \{5, 7\} \\ \midrule LightGBM & Number of estimators & \{100, 200, 300, 400, 500\} \\ LightGBM & Learning rate & \{0.01, 0.05, 0.1, 0.2\} \\ LightGBM & Maximum depth & \{5, 7\} \\ \midrule XGBoost & Number of estimators & \{100, 200, 300, 400, 500\} \\ XGBoost & Learning rate & \{0.01, 0.05, 0.1, 0.2\} \\ XGBoost & Maximum depth & \{5, 7\} \\ \midrule Chemprop & Depth & \{2, 3, 4\} \\ Chemprop & Message hidden dimension & \{300, 600, 1200\} \\ Chemprop & Dropout & $\mathcal{U}(0, 0.5)$ \\ Chemprop & Aggregation & \{\texttt{sum}, \texttt{mean}, \texttt{norm}\} \\ Chemprop & FFN hidden dimension & \{300, 600, 1200\} \\ Chemprop & Number of FFN layers & \{1, 2, 3\} \\ \midrule CheMeleon & FFN hidden dimension & \{300, 600, 1200\} \\ CheMeleon & Number of FFN layers & \{1, 2, 3\} \\ \midrule TabICLv2 & Number of estimators & \{8, 16, 32\} \\ TabICLv2 & Normalization method & \{\texttt{none}, \texttt{power}, \texttt{quantile}, \texttt{quantile\_rtdl}, \texttt{robust}\} \\ \midrule CLAMP & Probe model & \{\texttt{linear}, \texttt{nonlinear}\} \\ \bottomrule \end{tabular} \end{table}

\section{Results and Model evaluation}
\label{a:results}

\begin{figure}[ht]
    \centering
    \begin{subfigure}[b]{0.49\linewidth}
        \centering
        \includegraphics[width=\linewidth]{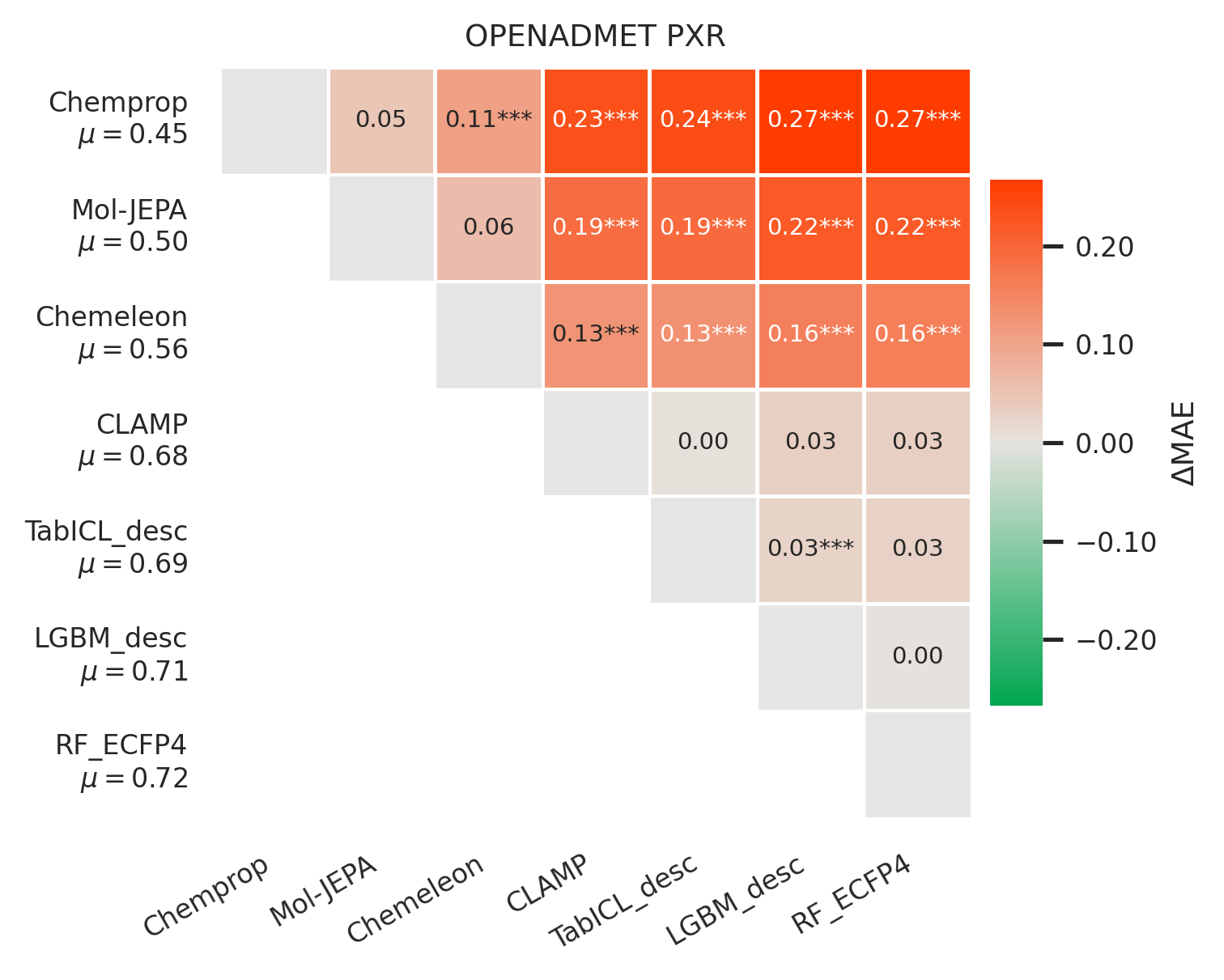}
        \caption{Pairwise statistical comparison using Wilcoxon test.}
        \label{sig1}
    \end{subfigure} 
    \hfill
    \begin{subfigure}[b]{0.49\linewidth}
        \centering
        \includegraphics[width=\linewidth]{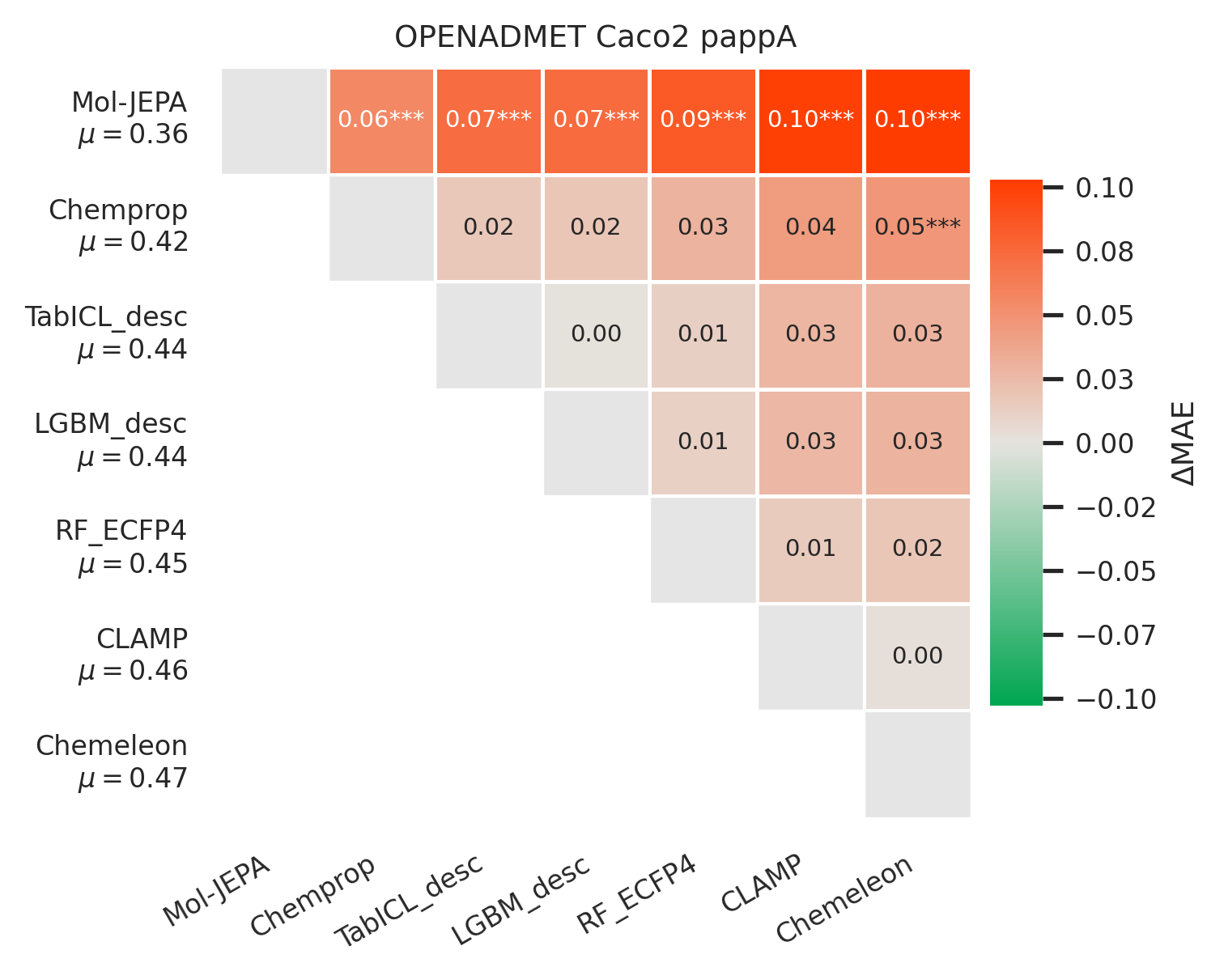}
        \caption{Pairwise statistical comparison using Wilcoxon test. }
        \label{sig2}
    \end{subfigure}
    \hfill
\end{figure}

\paragraph{Additional results}
Table \ref{tab:benchmarks_r2} reports the benchmark results using $R^2$ as evaluation metric. Consistent with the previous findings, Mol-JEPA achieves the strongest overall performance on average across datasets. Figures \ref{sig1} and \ref{sig2} highlight two representative examples on the cluster splits, demonstrating Mol-JEPA’s ability to significantly outperform the evaluated baseline methods in predictive accuracy.

\begin{figure}[t]
        \centering
        \includegraphics[width=0.6\linewidth]{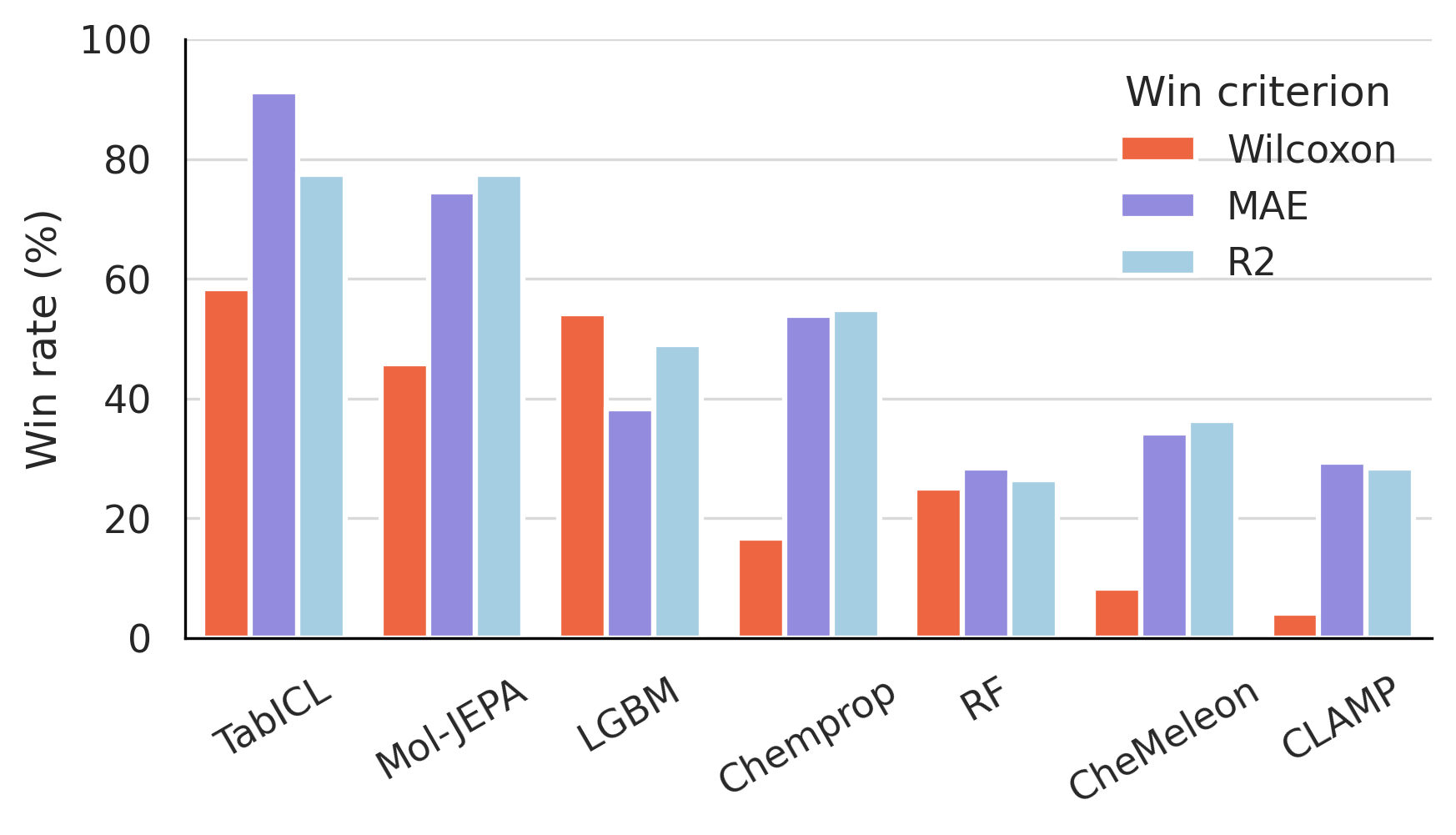}
        \caption{\textbf{Win rates on public competition splits}}
        \label{publicwin}
\end{figure}

We further evaluate model win rates on the public splits of ExpansionRX, ASAP, and PXR. These splits are generated using a temporal strategy, which, according to the competition organizers, better represents realistic drug discovery settings by preserving the chronological order of data generation. As shown in Figure \ref{publicwin}, TabICLv2 using AlvaDesc descriptors achieves higher Wilcoxon and MAE win rates than Mol-JEPA. However, Mol-JEPA delivers the strongest performance in terms of $R^2$. While these results indicate that there is still room for improvement through the incorporation of larger pretraining datasets and additional modalities, Mol-JEPA already demonstrates highly competitive performance. In particular, it consistently surpasses other foundation model approaches, which achieve substantially lower scores on both public and cluster-based splits.

\begin{table*} \centering \caption{$R^2$ comparison across benchmark datasets. Negative $R^2$ values are clipped to zero.} \label{tab:benchmarks_r2} \scriptsize \setlength{\tabcolsep}{2pt} \renewcommand{\arraystretch}{1.0} \begin{tabular}{l >{\columncolor{moljepa}}r >{\columncolor{moljepa}}r r r r r r r} \toprule \textbf{Dataset} & \makecell[c]{\textbf{Mol-JEPA}\\Best} & \makecell[c]{\textbf{Mol-JEPA}\\Embeddings} & \makecell[c]{\textbf{CLAMP}\\Nonlinear} & \makecell[c]{\textbf{CheMeleon}\\Finetuned} & \makecell[c]{\textbf{Chemprop}\\GNN} & \makecell[c]{\textbf{TabICLv2}\\AlvaDesc} & \makecell[c]{\textbf{RF}\\ECFP4} & \makecell[c]{\textbf{LGBM}\\AlvaDesc} \\ \midrule \rowcolor{slategray} \multicolumn{9}{l}{\textbf{ExpansionRx}}\\ Caco-2 Permeability & \textbf{0.43}\std{.06} & \underline{0.37}\std{.11} & 0.24\std{.18} & 0.29\std{.15} & 0.22\std{.20} & 0.14\std{.15} & 0.13\std{.07} & 0.14\std{.12} \\ Caco-2 Efflux & 0.30\std{.10} & 0.22\std{.16} & 0.06\std{.24} & 0.16\std{.12} & 0.23\std{.11} & \textbf{0.38}\std{.09} & 0.08\std{.03} & \underline{0.37}\std{.10} \\ LogD & \textbf{0.86}\std{.01} & 0.74\std{.07} & 0.69\std{.09} & \underline{0.85}\std{.01} & 0.74\std{.03} & 0.83\std{.06} & 0.41\std{.09} & 0.79\std{.05} \\ KSOL & \underline{0.51}\std{.06} & 0.38\std{.11} & 0.41\std{.07} & \textbf{0.53}\std{.08} & 0.19\std{.26} & \underline{0.51}\std{.06} & 0.27\std{.05} & 0.47\std{.04} \\ HLM CLint & 0.29\std{.06} & \underline{0.18}\std{.10} & 0.00\std{.00} & \textbf{0.31}\std{.16} & 0.00\std{.00} & 0.29\std{.07} & 0.07\std{.09} & 0.30\std{.02} \\ MLM CLint & 0.31\std{.22} & 0.31\std{.25} & \textbf{0.42}\std{.18} & 0.35\std{.24} & \underline{0.40}\std{.31} & 0.26\std{.20} & 0.21\std{.32} & 0.37\std{.27} \\ MBPB & \underline{0.65}\std{.08} & 0.63\std{.14} & 0.64\std{.03} & 0.66\std{.11} & 0.60\std{.08} & \textbf{0.72}\std{.10} & 0.30\std{.14} & 0.56\std{.15} \\ MGMB & \underline{0.42}\std{.28} & 0.38\std{.22} & 0.30\std{.27} & \textbf{0.44}\std{.29} & 0.30\std{.27} & 0.35\std{.48} & 0.11\std{.22} & 0.35\std{.40} \\ MPPB & \textbf{0.62}\std{.05} & 0.54\std{.09} & 0.46\std{.05} & 0.48\std{.15} & 0.41\std{.18} & \textbf{0.62}\std{.04} & 0.15\std{.24} & 0.51\std{.11} \\ \midrule \rowcolor{slategray} \multicolumn{9}{l}{\textbf{ASAP}}\\ MERS-CoV-2 Potency & 0.07\std{.12} & \textbf{0.19}\std{.24} & 0.00\std{.00} & 0.00\std{.00} & 0.00\std{.00} & \underline{0.14}\std{.12} & 0.03\std{.14} & \underline{0.14}\std{.03} \\ SARS-CoV-2 Potency & \underline{0.19}\std{.26} & 0.12\std{.23} & 0.00\std{.00} & 0.00\std{.00} & 0.00\std{.00} & \textbf{0.20}\std{.35} & 0.00\std{.00} & 0.13\std{.38} \\ LogD & \underline{0.57}\std{.08} & 0.53\std{.14} & 0.00\std{.00} & 0.31\std{.23} & 0.36\std{.23} & \textbf{0.64}\std{.14} & 0.13\std{.17} & 0.52\std{.13} \\ KSOL & \textbf{0.16}\std{.35} & \textbf{0.16}\std{.52} & 0.00\std{.00} & 0.00\std{.00} & 0.00\std{.00} & 0.00\std{.00} & 0.00\std{.00} & 0.00\std{.00} \\ HLM & \textbf{0.18}\std{.38} & \textbf{0.18}\std{.38} & 0.00\std{.00} & \underline{0.11}\std{.35} & 0.00\std{.00} & \underline{0.11}\std{.35} & 0.08\std{.17} & 0.09\std{.20} \\ MLM & 0.00\std{.00} & 0.00\std{.00} & 0.00\std{.00} & 0.00\std{.00} & 0.00\std{.00} & 0.00\std{.00} & \textbf{0.10}\std{.13} & 0.00\std{.00} \\ MDR1 Efflux & \textbf{0.11}\std{.09} & 0.09\std{.25} & 0.00\std{.00} & 0.00\std{.00} & 0.00\std{.00} & \underline{0.10}\std{.11} & 0.00\std{.00} & 0.00\std{.00} \\ \midrule \rowcolor{slategray} \multicolumn{9}{l}{\textbf{PXR}}\\ PXR Activity & \textbf{0.66}\std{.08} & \underline{0.63}\std{.12} & 0.41\std{.13} & 0.57\std{.23} & 0.52\std{.17} & 0.20\std{.43} & 0.31\std{.26} & 0.22\std{.39} \\ \midrule \rowcolor{slategray} \multicolumn{9}{l}{\textbf{Biogen ADME}}\\ Solubility & \textbf{0.50}\std{.06} & \underline{0.44}\std{.07} & 0.33\std{.05} & 0.33\std{.09} & 0.27\std{.15} & 0.44\std{.01} & 0.18\std{.08} & 0.41\std{.06} \\ HLM CLint & \textbf{0.54}\std{.09} & 0.50\std{.14} & 0.33\std{.06} & 0.51\std{.13} & 0.44\std{.09} & \textbf{0.54}\std{.06} & 0.12\std{.12} & \textbf{0.54}\std{.11} \\ RLM CLint & \underline{0.58}\std{.09} & 0.57\std{.06} & 0.42\std{.00} & 0.57\std{.07} & 0.49\std{.07} & \underline{0.58}\std{.06} & 0.15\std{.11} & 0.57\std{.08} \\ HPPB & \textbf{0.73}\std{.07} & \underline{0.68}\std{.07} & 0.51\std{.12} & 0.51\std{.08} & 0.47\std{.08} & 0.68\std{.09} & 0.12\std{.17} & 0.57\std{.14} \\ RPPB & \underline{0.43}\std{.27} & 0.41\std{.20} & \textbf{0.47}\std{.21} & 0.27\std{.24} & 0.20\std{.23} & 0.28\std{.40} & 0.00\std{.00} & 0.14\std{.42} \\ MDR1 Efflux & \textbf{0.60}\std{.07} & \underline{0.56}\std{.10} & 0.43\std{.07} & 0.55\std{.11} & 0.53\std{.15} & 0.53\std{.21} & 0.12\std{.12} & 0.51\std{.24} \\ \midrule \rowcolor{slategray} \textbf{Average $R^2$} & \textbf{0.42} & \underline{0.38} & 0.27 & 0.34 & 0.28 & 0.37 & 0.13 & 0.33 \\ \bottomrule \end{tabular} \end{table*}

\begin{figure}[t]
    \centering
    \begin{subfigure}[b]{0.50\linewidth}
        \centering
        \includegraphics[width=\linewidth]{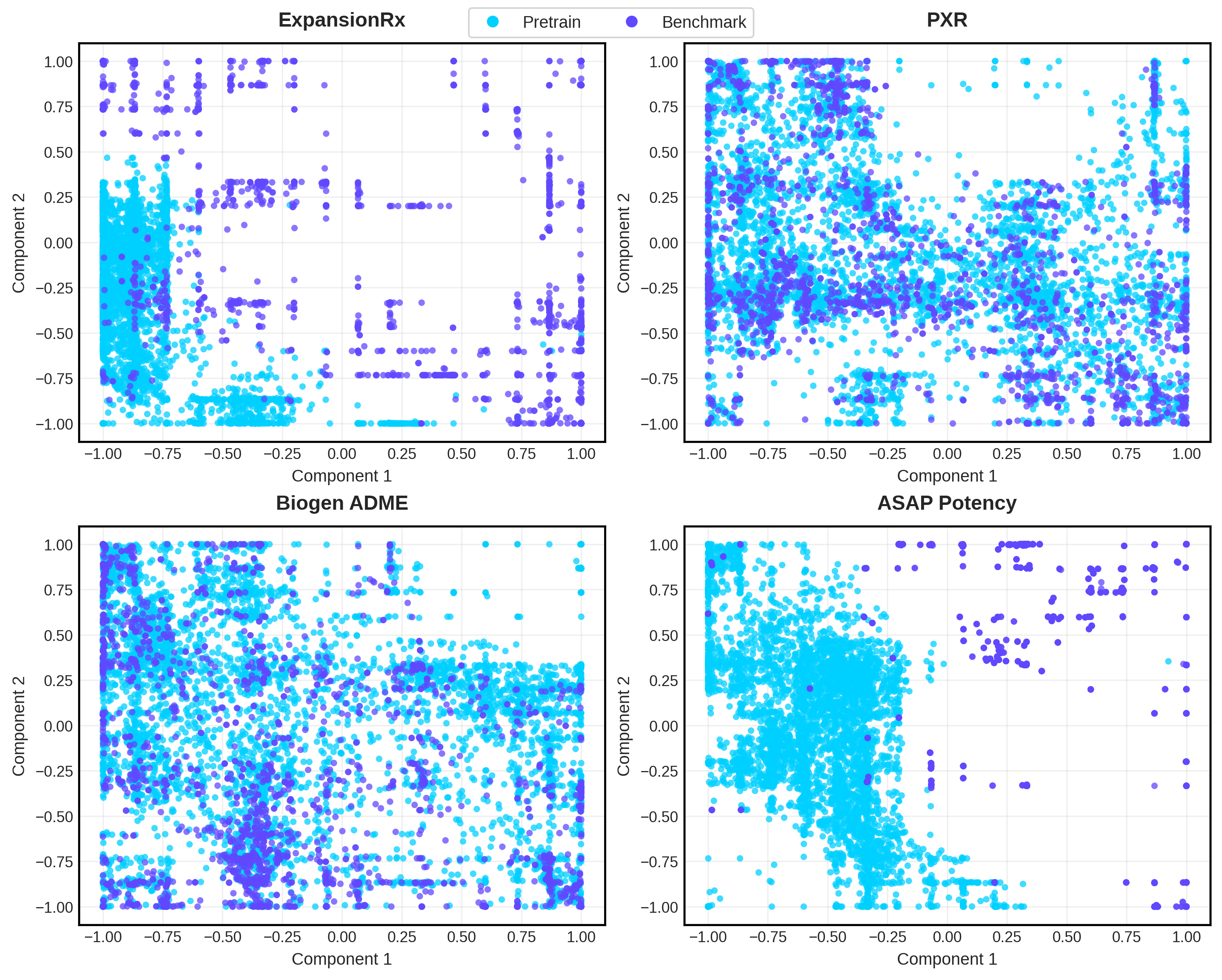}
        \caption{\textbf{Generative Topographical Mappings} for different benchmark families and pretraining data.}
        \label{gtm}
    \end{subfigure} 
    \hfill
    \begin{subfigure}[b]{0.45\linewidth}
        \centering
        \includegraphics[width=\linewidth]{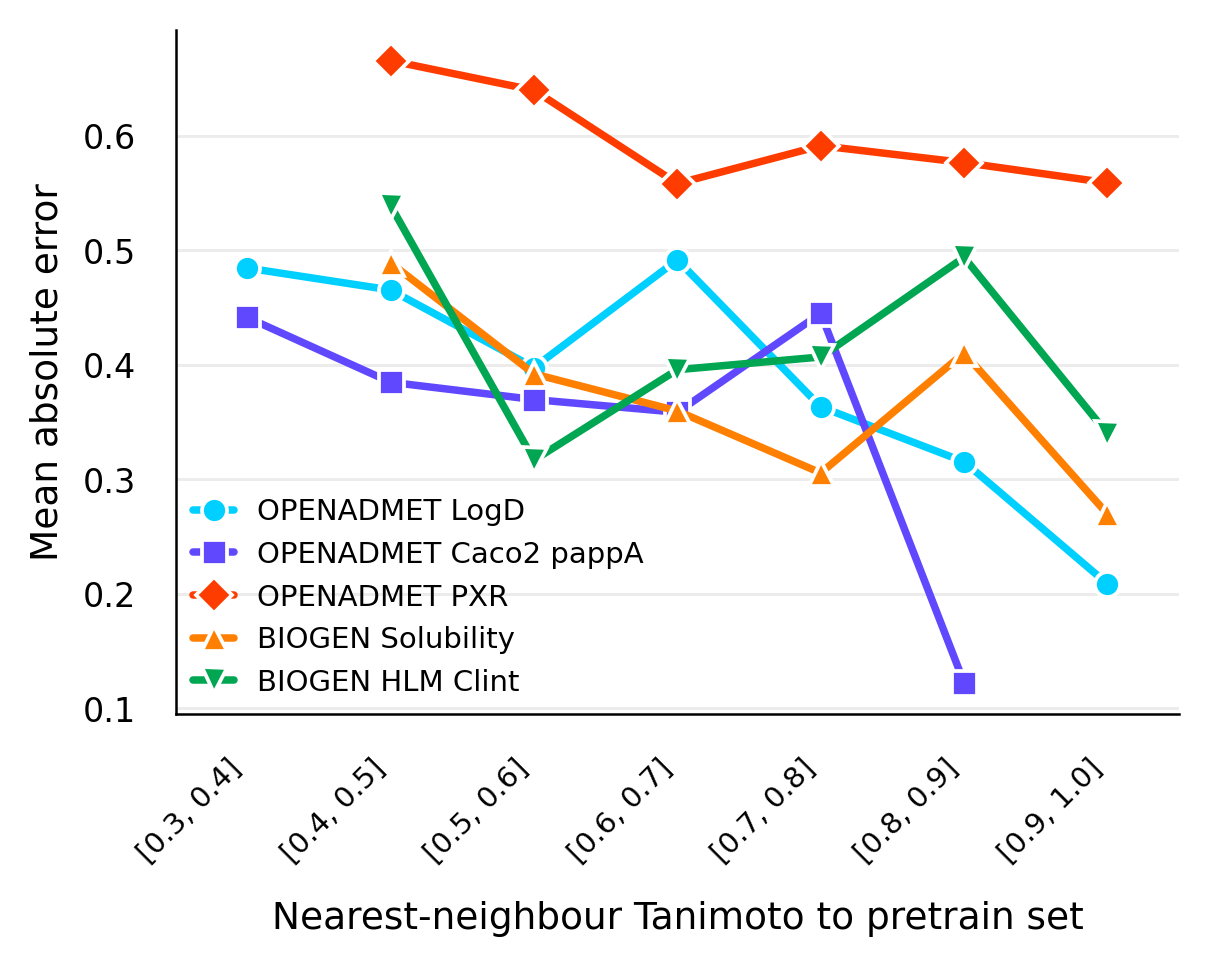}
        \caption{\textbf{Pretraining data coverage determines downstream performance}. }
        \label{pretrainerro}
    \end{subfigure}
    \hfill
    \caption{Pretraining data analysis}
\end{figure}

\paragraph{Molecular space coverage}
To understand the role of the pretraining data for the downstream tasks, we analyze the molecular similarity using generative topographic mappings (GTM) \cite{bishop1998gtm}. We randomly sample 5,000 molecules from the pretraining corpus and 2,000 molecules from the benchmark datasets, convert them into ECFP4 fingerprints, and project them into a shared GTM space. The resulting visualization, shown in Figure \ref{gtm}, reveals that while some benchmark datasets are well represented within the pretraining distribution, others occupy more distant regions of chemical space. Notably, Mol-JEPA tends to achieve stronger performance on benchmark families with more pretraining data coverage, suggesting that representation of relevant chemical space during pretraining contributes to downstream predictive accuracy. To further investigate this relationship, we selected five datasets and computed, for each molecule, its distance to the full pretraining corpus using ECFP4 fingerprints. Figure \ref{pretrainerro} illustrates the relationship between prediction error and binned molecular similarity. The results show a clear trend of increasing Mol-JEPA prediction error with increasing distance from the pretraining data, indicating that molecules located further from the pretraining distribution are more challenging to predict accurately. These findings suggest that expanding the size and diversity of the pretraining corpus could further improve model performance by increasing coverage of relevant chemical space.

\section{Pretraining data scaling}
\begin{figure}[t]
    \centering
    \begin{subfigure}[b]{0.49\linewidth}
        \centering
        \includegraphics[width=\linewidth]{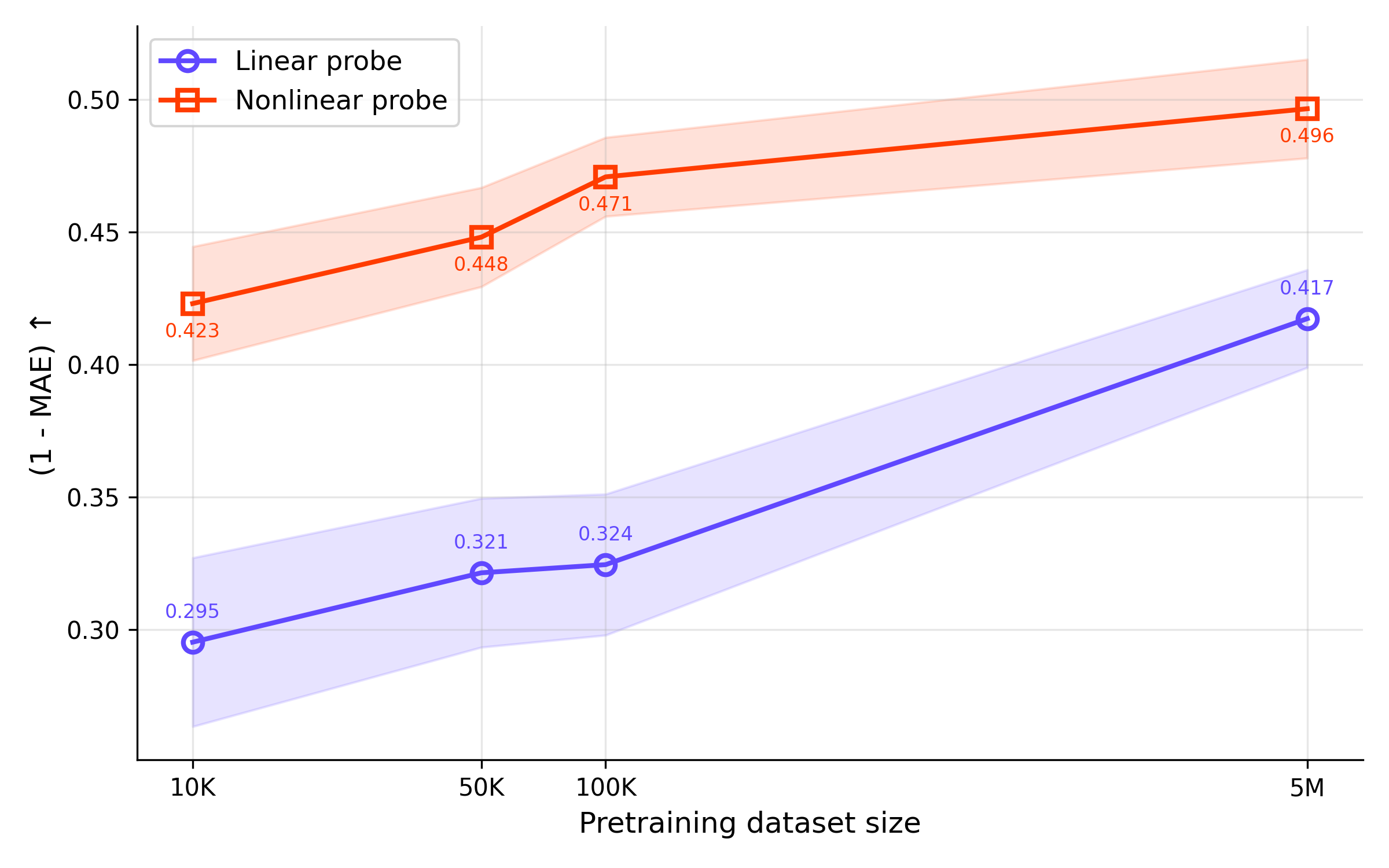}
        \caption{\textbf{Effect of pretraining dataset size on performance.}}
        \label{pretrain_size}
        \end{subfigure}
    \hfill
    \begin{subfigure}[b]{0.49\linewidth}
        \centering
        \includegraphics[width=\linewidth]{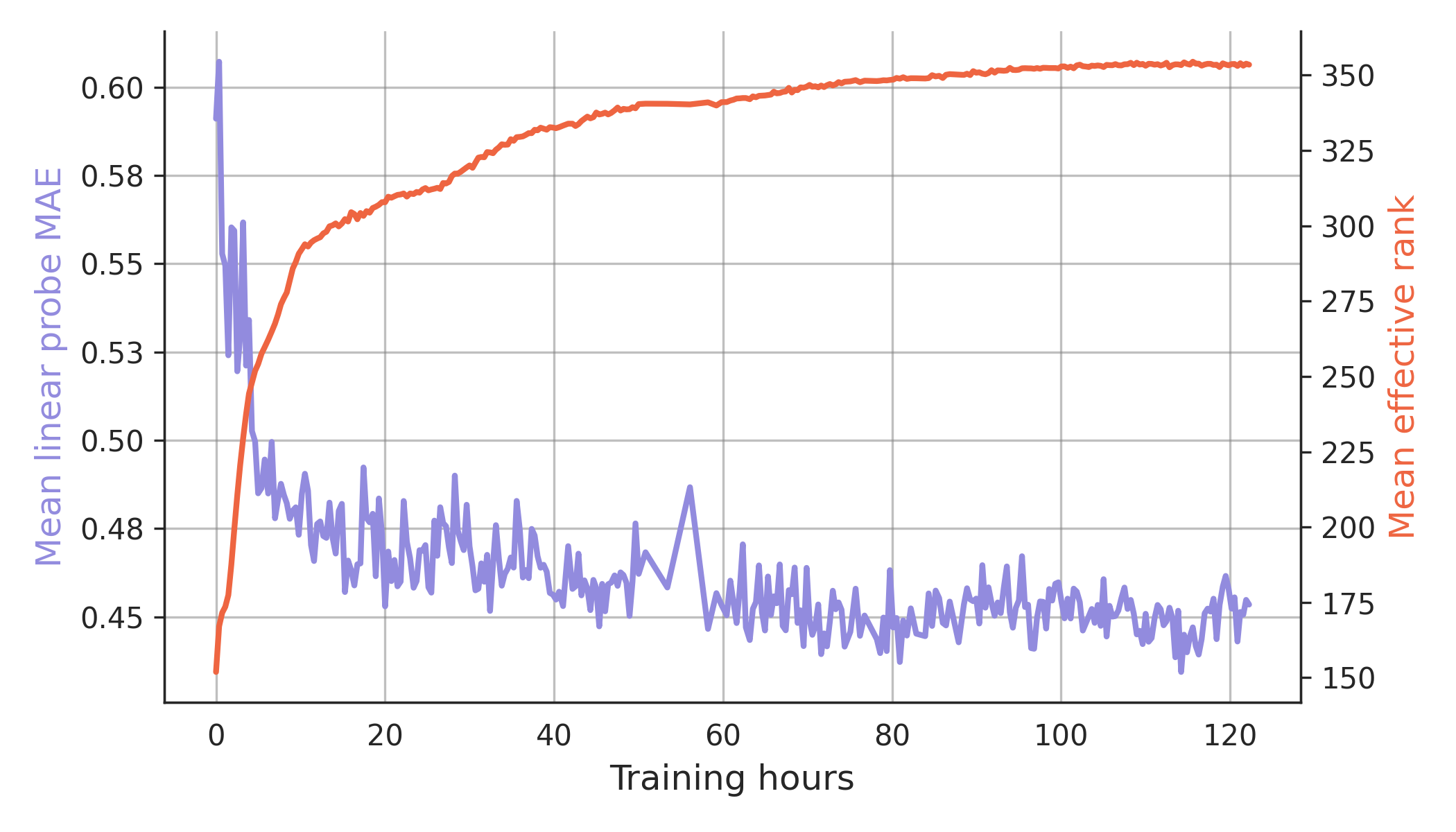}
        \caption{\textbf{Effective rank and downstream performance.}}
    \end{subfigure} 
    \hfill
    \caption{Training configurations}
    \label{fig:effects}
\end{figure}

We analyze the effect of pretraining dataset size on downstream performance. As shown in Figure \ref{pretrain_size}, performance steadily improves with increasing pretraining data for both linear and nonlinear probes across all downstream datasets (reflected by the confidence bands). To better highlight this trend, performance is reported as 1-MAE, where higher values indicate better performance. The observed scaling behavior suggests that Mol-JEPA benefits from additional pretraining data and further increases in dataset size are expected to yield additional gains in downstream performance. 

\section{Ablation studies}
\label{a:ablations}
We perform several experiments to assess the sensitivity of different parameter and architectural choices, which are summarized in Table \ref{tab:ablations}. For computational reasons, the experiments are conducted using a subset of 100,000 randomly sampled data points. 

\paragraph{Experimental modalities.} Overall, we find that the performance differences introduced by the experimental modalities are moderate. This is likely because the modality data contains many missing values, imbalanced distributions, and potentially noisy measurements, limiting the amount of useful information that can be extracted. To better understand how this information should be incorporated, we conduct two additional ablation studies. First, we investigate whether modality information should be used as labels in a semi-supervised setting or incorporated through latent prediction within the JEPA framework. As summarized in Table \ref{tab:ablations}, the semi-supervised alternative consistently underperforms JEPA across both linear and non-linear probing settings, increasing the MAE from 0.436 to 0.467 and from 0.393 to 0.417, respectively. These results suggest that JEPA more effectively exploits the available modality information, supporting the hypothesis that learning predictive latent representations is a more robust strategy for integrating noisy and incomplete experimental data than directly predicting modality labels. Second, we assess the impact of removing the experimental modalities entirely. In this setting, downstream performance consistently deteriorates, demonstrating that the modalities still provide valuable information that improves the quality and transferability of the learned molecular representations (MAE 0.436 vs. 0.449 and 0.393 vs. 0.407).

\begin{table}[ht]
  \centering
  \caption{Model architecture and training ablations on a subset of 100k. The reported numbers are the best downstream MAEs after training up to 1000 epochs, averaged across all benchmark datasets.}
  \label{tab:ablations}
  \setlength{\tabcolsep}{8pt}
  \renewcommand{\arraystretch}{1.15}
  \begin{tabular}{llcc}
    \toprule
    \textbf{Group} & \textbf{Variant} & \textbf{Linear probes mean} & \textbf{Nonlinear probes mean} \\
    \midrule

    \rowcolor{groupbg}
    \multicolumn{4}{l}{Modalities} \\
    & All modalities & \textbf{0.436} & \textbf{0.393} \\
    & Semi-Supervised    & 0.467 & 0.417  \\
    & No experimental  & 0.449 & 0.407 \\
    \midrule
    
    \rowcolor{groupbg}
    \multicolumn{4}{l}{Regularization} \\
    & SIGReg on targets & \textbf{0.436} & \textbf{0.393}  \\
    & SIGReg on CLS      & 0.445 & 0.406  \\
    \midrule

    \rowcolor{groupbg}
    \multicolumn{4}{l}{Loss projection} \\
    & W/o projection    & \textbf{0.436} & \textbf{0.393}  \\
    & W/ projection    & 0.468 & 0.488  \\
    \midrule
    
    \rowcolor{groupbg}
    \multicolumn{4}{l}{Embedding dimension} \\
    & 256               & 0.483 & 0.465  \\
    & 512              & 0.436 & \textbf{0.393}  \\
    & 1024              & \textbf{0.435} & 0.397  \\
    \bottomrule
  \end{tabular}
\end{table}

\paragraph{Regularization}
There are several ways to apply the SIGReg loss within the JEPA framework. First, isotropy can be enforced on the CLS embeddings. Second, the loss can be applied directly to the predicted embeddings. Third, it can be applied to the target embeddings. We observe training collapse when jointly optimizing the prediction objective and the SIGReg regularization on the predicted embeddings. In contrast, applying the regularization to the target embeddings yields stable training and the best downstream performance. These results suggest that regularizing the target representations encourages a well-structured latent space while allowing the predictor to focus on the alignment objective.

\paragraph{Loss projection}
Similar to contrastive learning approaches such as SimCLR, we investigate whether applying the training objective in a dedicated projection space is beneficial. As shown in Table \ref{tab:ablations}, removing the projection head leads to improved downstream performance for both the linear probe (0.468 vs. 0.436) and the non-linear probe (0.488 vs. 0.393). These results suggest that, unlike in contrastive learning, introducing a separate projection space is not advantageous in our JEPA-based setting. A possible explanation is that the JEPA objective already promotes informative latent representations, making an additional projection head unnecessary and potentially causing the backbone representations to lose information relevant for downstream tasks.

\paragraph{Embedding dimension.} We further investigate the impact of the embedding dimension on downstream performance. As shown in Table \ref{tab:ablations}, increasing the embedding dimension from 256 to 512 substantially improves performance, reducing the MAE for both the linear probe (0.483 to 0.436) and the non-linear probe (0.465 to 0.393). Further increasing the dimension to 1024 yields nearly identical results (0.435 and 0.397, respectively), indicating diminishing returns from additional model capacity. Overall, these findings suggest that larger embedding dimensions are beneficial up to a certain point, after which performance largely saturates. Consequently, an embedding dimension of 512 appears to provide a favorable trade-off between representation capacity and downstream performance.

\begin{figure}[h]
        \centering
        \includegraphics[width=0.4\linewidth]{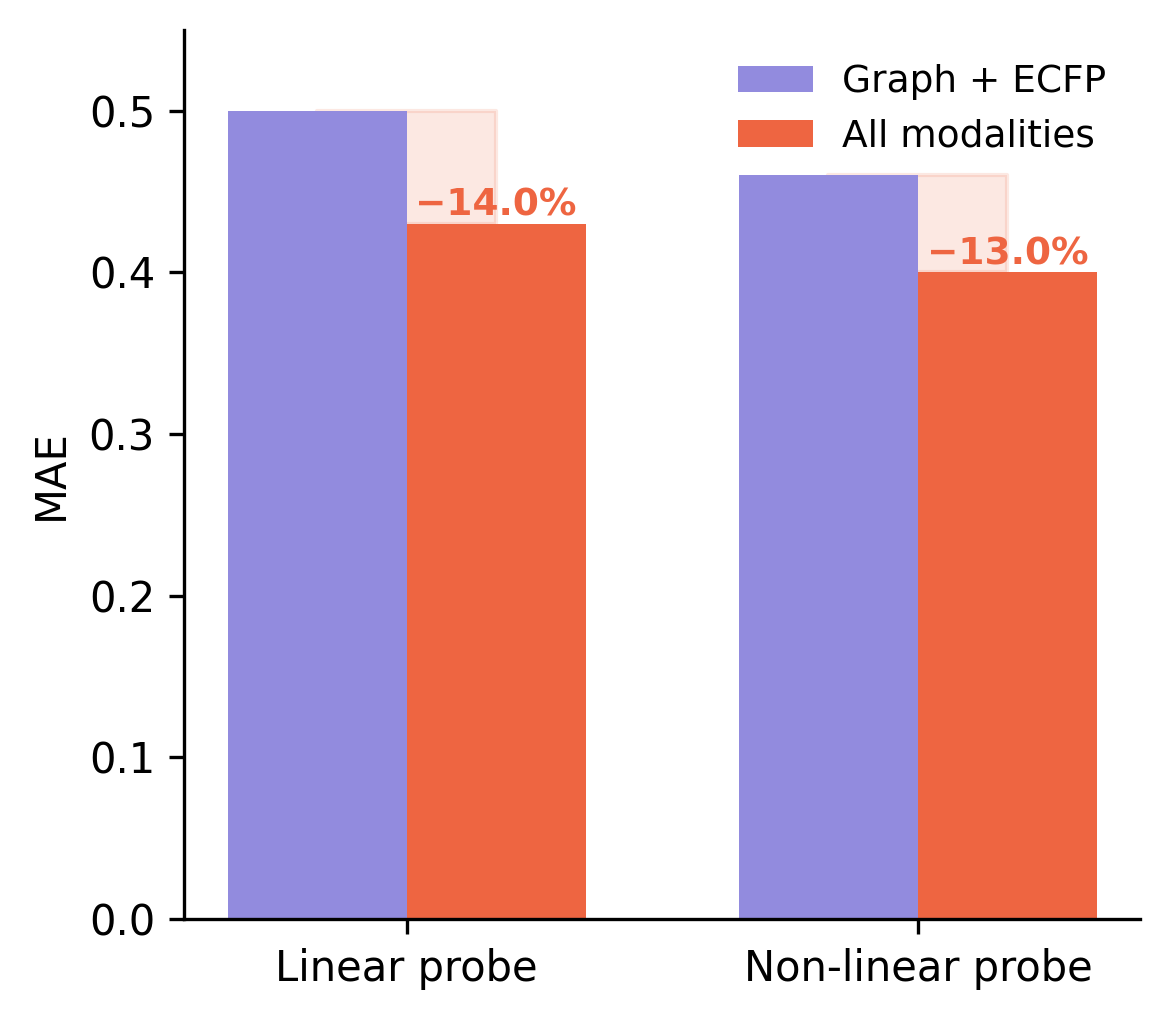}
        \caption{\textbf{Multimodal performance gains}. We re-train the model on the full dataset using only Graph and ECFP4 modalities. We find that the downstream performance is significantly better when using all modalities. }
        \label{modalitygain}
\end{figure}

\paragraph{Multimodal benefits}
To quantify the benefit of incorporating additional experimental modalities, we perform a controlled comparison using the full pretraining dataset. Specifically, we train (1) a model using all available modalities and (2) a model using only molecular graphs and ECFP4 fingerprints, which can be computed directly from molecular structure. We evaluate both models on the downstream benchmark datasets using linear and non-linear probe models based on the CLS representation. As shown in Figure \ref{modalitygain}, we find that incorporating the additional modalities reduces the MAE by 14\% for the linear probe and 13\% for the non-linear probe. These results demonstrate that modalities provide valuable complementary information that significantly improves the quality of the learned molecular representations.

\section{Representational similarity}

\begin{figure}
        \centering
        \includegraphics[width=0.7\linewidth]{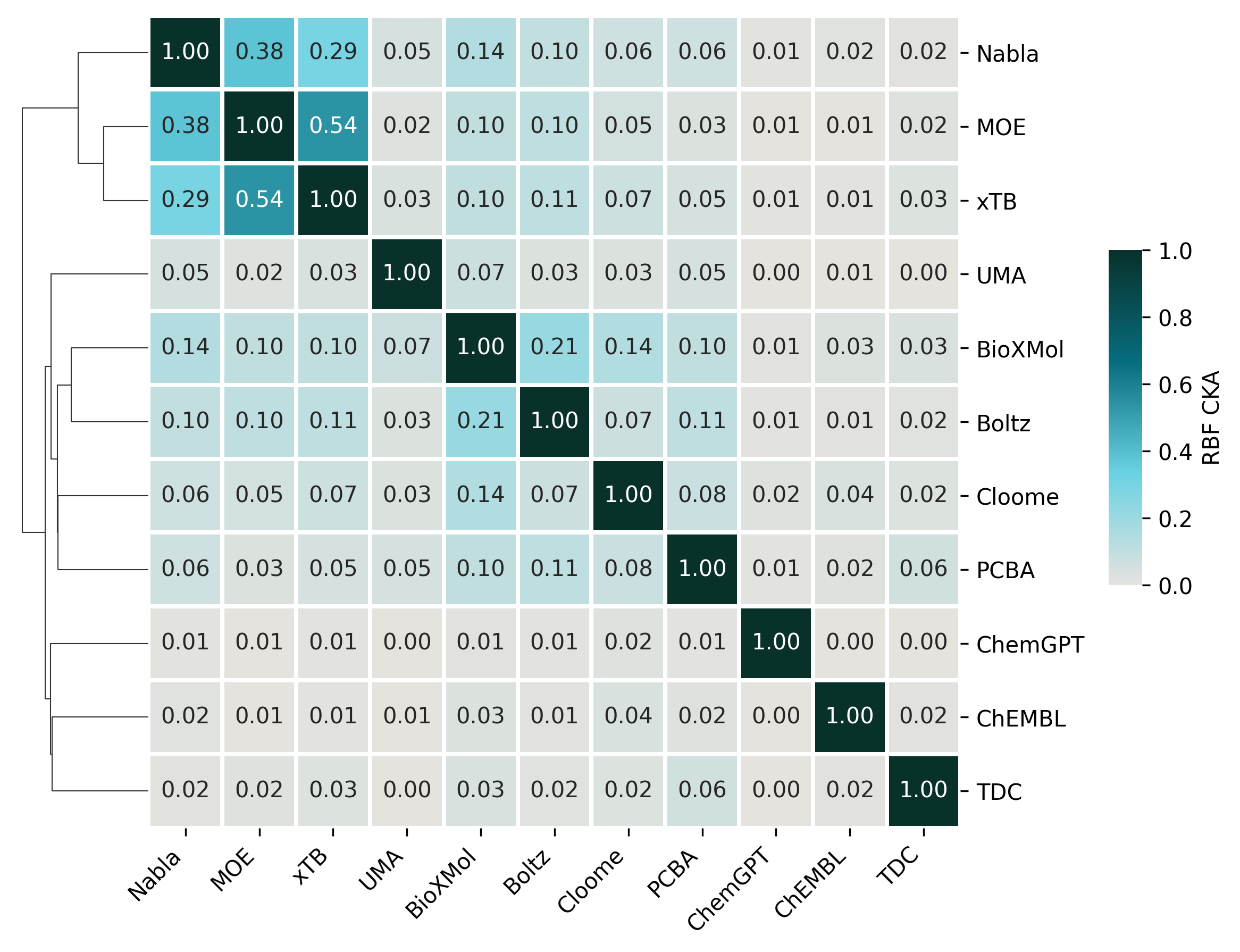}
        \caption{Centered Kernel Alignment}
        \label{cka}
\end{figure}

\paragraph{Representation similarity analysis.} Figure \ref{cka} shows the pairwise Centered Kernel Alignment (CKA) similarities between modality-specific embeddings. Clear clusters emerge among modalities that capture related information. Most notably, the quantum chemistry modalities, including Nabla, MOE, and xTB, exhibit the highest mutual similarities, reflecting their shared focus on physicochemical and electronic properties. The cellular profiling modalities CLOOME and BioXMol form a distinct cluster, which is expected given their overlap in underlying datasets and biological measurements. Interestingly, BioXMol also shows similarity with Boltz, suggesting that cellular phenotypic profiles may implicitly encode information related to protein binding interactions. In contrast, ChEMBL and TDC show comparatively low similarity to most other modalities, indicating that they contribute complementary information not captured by the remaining data sources. Overall, these results suggest that Mol-JEPA organizes modality-specific representations according to their underlying biological and physicochemical relationships while preserving information unique to individual modalities.

\section{In-context learning comparison}
\begin{figure}[t] \centering \includegraphics[width=0.95\linewidth]{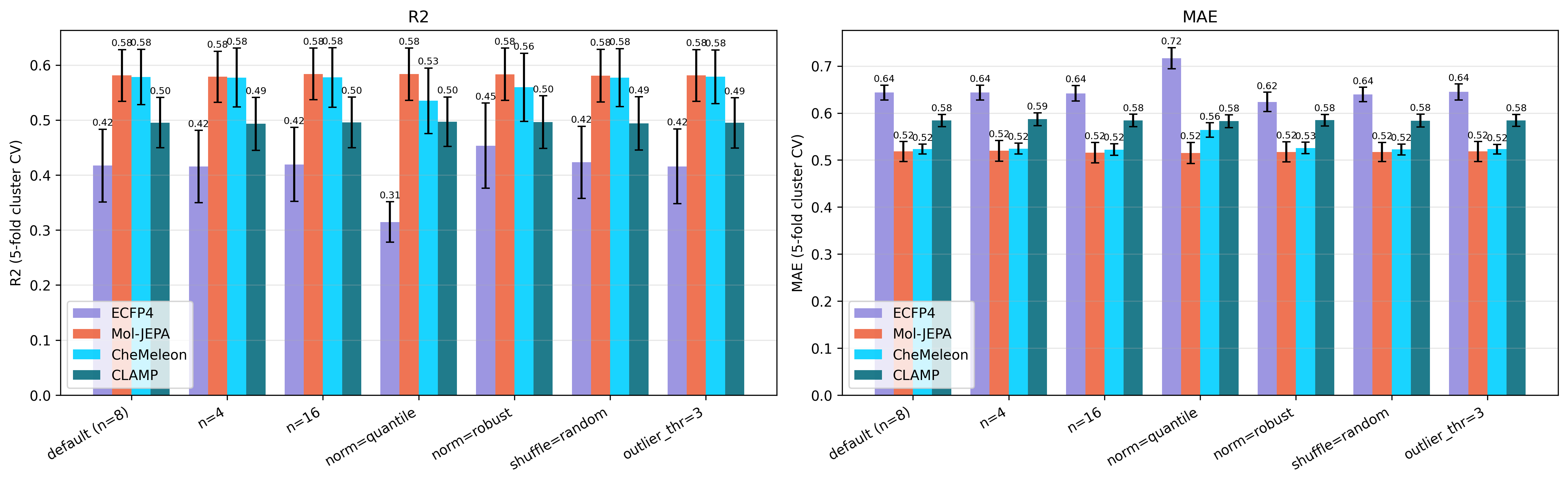} \caption{OpenADMET PXR TabICLv2 comparison.} \label{fig:tabicl_pxr} \end{figure}

\begin{figure}[t] \centering \includegraphics[width=0.95\linewidth]{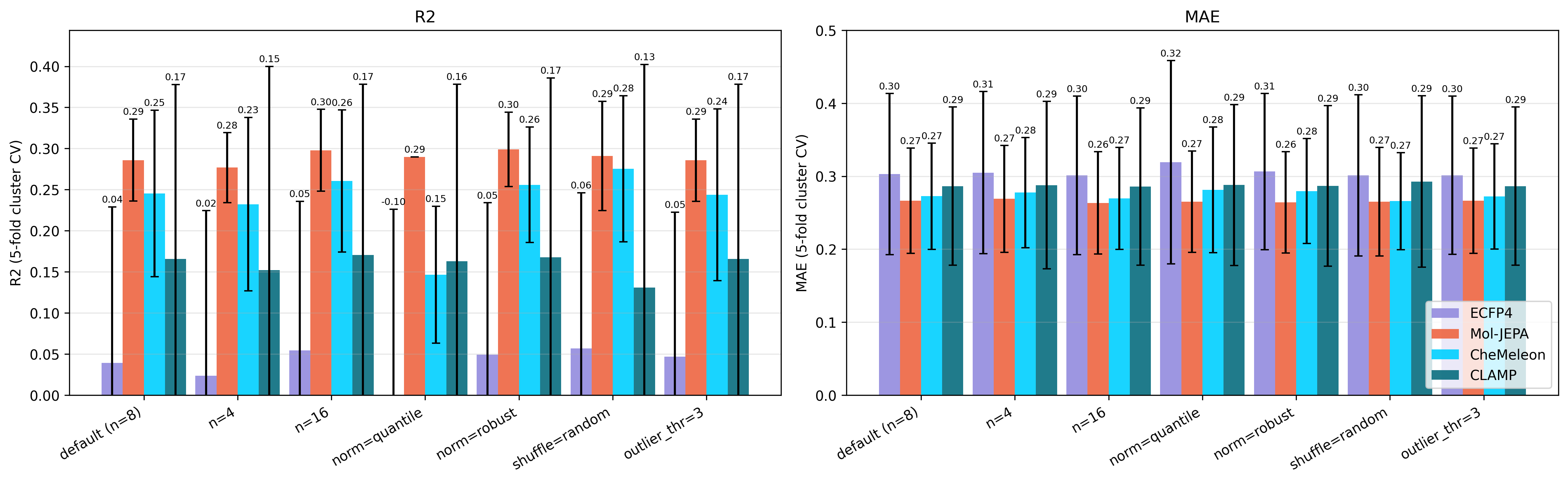} \caption{OpenADMET Caco2 Efflux TabICLv2 comparison.} \label{fig:tabicl_caco} \end{figure}

\begin{figure}[t] \centering \includegraphics[width=0.95\linewidth]{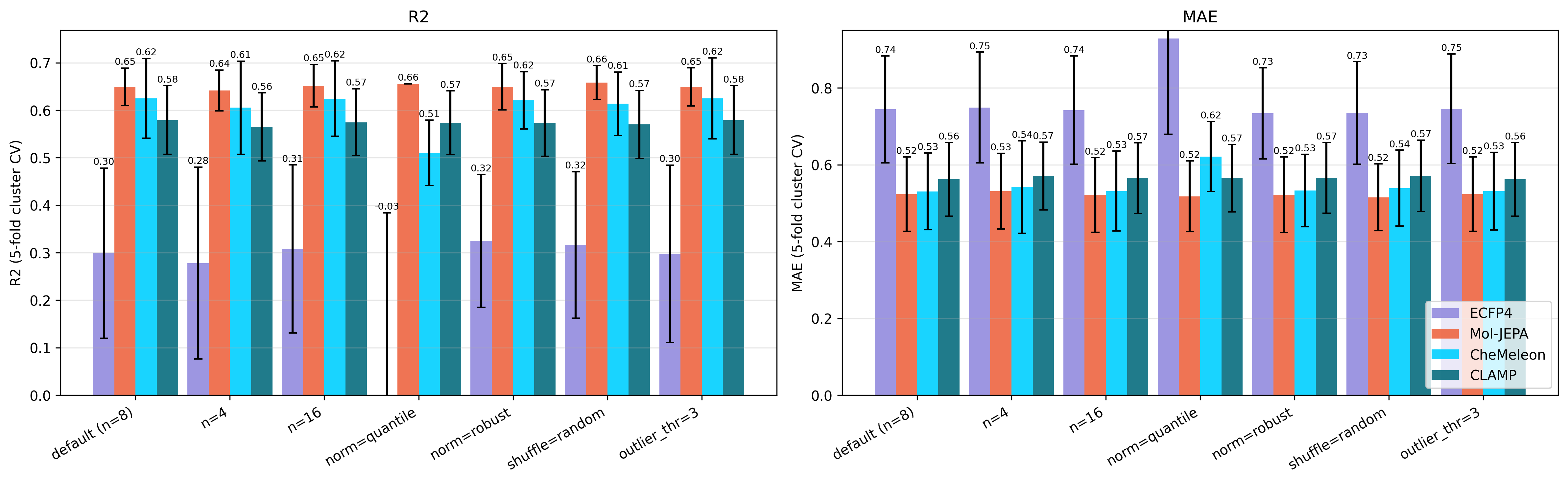} \caption{OpenADMET LogD TabICLv2 comparison.} \label{fig:tabicl_logd} \end{figure}

We assess the effectiveness of Mol-JEPA representations for in-context learning by comparing them with widely used molecular representations. As illustrated in Figures~\ref{fig:tabicl_pxr}-\ref{fig:tabicl_logd}, Mol-JEPA CLS-token embeddings achieve superior performance across most tasks for different TabICLv2 hyperparameters. This finding indicates that the multimodal architecture effectively combines complementary information from different molecular modalities, resulting in richer representations that are particularly beneficial for in-context learning.

\end{document}